# Lagged backward-compatible physics-informed neural networks for unsaturated soil consolidation analysis

Dong Li[1], Shuai Huang[2], Yapeng Cao[3*], Yujun Cui[4], Xiaobin Wei[5], Hongtao Cao[6]

**Abstract:** This study develops a Lagged Backward-Compatible Physics-Informed Neural Network (LBC-PINN) for simulating and inverting one-dimensional unsaturated soil consolidation under long-term loading. To address the challenges of coupled air–water pressure dissipation across multi-scale time domains, the framework integrates logarithmic time segmentation, lagged compatibility loss enforcement, and segment-wise transfer learning. In forward analysis, the LBC-PINN with recommended segmentation schemes accurately predicts pore-air and pore-water pressure evolutions, which are validated against finite element method (FEM) results with mean absolute errors below $10^{-2}$ across time up to $10^{10}$ seconds. A simplified segmentation strategy based on the characteristic air-phase dissipation time improves the computational efficiency while preserving the predictive accuracy. Sensitivity analyses confirm the framework's robustness across air-to-water permeability ratios $k_a/k_w$ from $10^{-3}$ to $10^3$.

[1] PhD, Department of Civil, Environmental, and Infrastructure Engineering, George Mason University, Fairfax, VA 22030, USA
[2] PhD, National Institute of Natural Hazards, Ministry of Emergency Management, Beijing 100085, China
[3*] PhD, State Key Laboratory of Cryospheric Science and Frozen Soil Engineering, Northwest Institute of Eco-Environment and Resources, Chinese Academy of Sciences, Lanzhou 730000, China; Navier Laboratory, École Nationale des Ponts et Chaussées, 77455 Marne-la-Vallée Cedex 2, France (Corresponding author) email: yapeng.cao@enpc.fr
[4] Professor, Navier Laboratory, École Nationale des Ponts et Chaussées, 77455 Marne-la-Vallée Cedex 2, France
[5] PhD, School of Civil Engineering, Hebei University of Engineering, Handan 056038, China
[6] PhD, College of Civil Engineering, Zhejiang University of Technology, Hangzhou 310023, China

# Notation

| Symbol | Description | Unit |
|---|---|---|
| $C_a$ | Interactive constant associated with the air phase | – |
| $C_w$ | Interactive constant associated with the water phase | – |
| $c_v^a$ | Coefficient of consolidation with respect to the air phase | m²/s |
| $c_v^w$ | Coefficient of consolidation with respect to the water phase | m²/s |
| $H$ | Thickness of soil layer | m |
| $k_a$ | Air permeability coefficients | m/s |
| $k_w$ | Water permeability coefficients | m/s |
| $k_a/k_w$ | Air-to-water permeability ratio | – |
| $\mathcal{L}_{\mathrm{BC}}$ | Boundary condition loss term | – |
| $\mathcal{L}_{\mathrm{IC}}$ | Initial condition loss term | – |
| $\mathcal{L}_R$ | PDE residual (interior) loss term | – |
| $\mathcal{L}_S$ | Lagged compatibility loss term | – |
| $\mathcal{L}_{\mathrm{STD}}$ | Total loss of the standard PINN | – |
| $m_1^a$ | Coefficients of air volume change with respect to the change in the net normal stress | /kPa |
| $m_2^a$ | Coefficients of air volume change with respect to the change in the net matric suction | /kPa |
| $m_1^w$ | Coefficients of water volume change with respect to the change in the net normal stress | /kPa |
| $m_2^w$ | Coefficients of water volume change with respect to the change in the net matric suction | /kPa |
| $MAE$ | Mean absolute error | kPa |
| $MRE$ | Mean relative error | – |
| $n$ | Porosity of soil | – |
| $R$ | Universal air constant | J/mol/K |
| $R^2$ | Coefficient of determination | – |
| $S_r$ | Degree of saturation | % |
| $T$ | Absolute temperature | K |
| $T_{\max}$ | Maximum simulation / training time horizon | s |
| $t$ | Time | s |
| $t_s$ | Characteristic air-phase dissipation time | s |
| $u_a(z,t)$ | Excess pore-air pressure | kPa |
| $u_w(z,t)$ | Excess pore-water pressure | kPa |
| $u_a^0$ | Initial pore-air pressure | kPa |
| $u_w^0$ | Initial pore-water pressure | kPa |
| $z$ | Vertical coordinate (depth, measured from ground surface) | m |
| $\gamma_w$ | Unit weight of water | kN/m³ |
| $\rho_a$ | Density of air | kg/m³ |
| $\rho_w$ | Density of water | kg/m³ |
| $\sigma$ | Total vertical stress | kPa |
| $\sigma'$ | Effective vertical stress | kPa |

| $\theta$ | Vector of trainable neural-network parameters (weights and biases) | – |
| $\omega_{BC}$ | Loss weight parameter for boundary condition loss term | – |
| $\omega_{IC}$ | Loss weight parameter for initial condition loss term | – |
| $\omega_R$ | Loss weight parameter for PDE residual loss term | – |
| $\omega_s$ | Loss weight parameter for lagged compatibility loss term | – |

## 1. Introduction

Consolidation in geotechnical engineering refers to the dissipation of excess pore-water pressure in soils under external loading, resulting in time-dependent settlement (Biot, 1941; Esrig, 1968; Mesri and Rokhsar, 1974; Cargill, 1984; Conte, 2004; Huang et al., 2010). The time-dependent settlement behavior of saturated soils is traditionally modeled using Terzaghi's one-dimensional consolidation theory, which formulates consolidation as the dissipation of excess pore-water pressure induced by sustained loading based on idealized assumptions such as constant permeability and compressibility of soil (Terzaghi, 1943; Mesri and Rokhsar, 1974). In many practical situations, such as compacted fills, natural deposits above the water table, or structures subjected to fluctuating groundwater levels, soils remain partially unsaturated under sustained loading and environmental fluctuations (Barden, 1965; Lloret and Alonso, 1980; Ho et al., 2014; Ho and Fatahi, 2016). When the air phase is present and interacts with the water phase, the governing mechanisms of consolidation differ from those in saturated soils (Conte, 2004). Early studies on unsaturated soil consolidation include Scott's (1963) estimation for soils with occluded air bubbles, Biot's (1941) general consolidation theory applicable to such conditions, and Barden's (1965) one-dimensional consolidation analysis of compacted unsaturated clay.

Assuming that the air and water phases are continuous, Fredlund and Hasan (1979) further developed a one-dimensional consolidation theory for unsaturated soils, formulating coupled governing equations for water and air phases based on mass conservation and Darcy's law. Traditionally, these behaviors are described and analyzed using ordinary and partial differential equations (ODEs and PDEs), which form the basis for analytical and numerical solutions in unsaturated soil consolidation analysis (Wang et al., 2020;

Zhou et al., 2017; Briaud, 2023). Several analytical solutions have been developed for unsaturated soil consolidation under one-way (single-drainage) and two-way (double-drainage) conditions, accounting for air-water coupling, including those by Qin et al. (2008, 2010), Zhou et al. (2014), Zhou et al. (2017), Shan et al. (2012), Ho et al. (2014), Ho and Fatahi (2016, 2018). Qin et al. (2008, 2010) developed a Laplace transformation-based method to predict one-dimensional settlement of unsaturated soil using the governing equations of Fredlund and Hasan (1979). Shan et al. (2012) obtained an exact solution for one-dimensional consolidation of unsaturated soils through the variables-separation method. Ho et al. (2014) derived an exact analytical solution for the governing equations of one-dimensional consolidation in an unsaturated soil stratum using eigenfunction expansion and Laplace transformation. However, deriving a closed-form solution typically necessitates advanced mathematical methods such as Laplace/inverse transforms and the Cayley-Hamilton theory, and requires laborious resolution of the associated eigenproblems. These algebraic derivations are cumbersome and constitute a principal computational bottleneck. Numerical techniques are therefore the preferred method to approximate PDEs, such as finite-element (FE) methods (Wong et al., 1998; Cheng et al., 2017; Tang et al., 2018; Liu et al., 2014), and differential quadrature (DQ) method (Zhou 2013; Zhou and Zhao, 2014). However, conventional numerical methods face several limitations, including susceptibility to mesh distortion under large deformations, convergence difficulties in strongly coupled analyses, high computational demands for inverse modeling, and challenges in representing complex boundary conditions for both air and water phases (Reddy, 1993).

Recent advances in scientific machine learning have introduced Physics-Informed Neural Networks (PINNs) as a mesh-free method for solving partial differential equations by embedding governing equations, boundary conditions, and initial conditions into the loss function of a neural network (Cuomo et al., 2022; Lawal et al., 2022). This approach was pioneered by Raissi et al. (2019) and applied by Bekele (2021) to Terzaghi's consolidation. Bekele (2021) demonstrated that a PINN can accurately reproduce the 1D consolidation solution (forward problems) and estimate the consolidation coefficient from measured data (inverse problem) by minimizing combined physics and data loss. Wang et al. (2024) further developed a novel PINN for forward and inverse analyses of two-dimensional (2D) soil consolidation using limited site-specific measurements with special attention to the out-of-sample prediction performance. Lan et al. (2024) enhanced physics-informed neural networks (PINNs) by incorporating hard constraints (PINNs-H) to model excess pore-water pressure variation in nonlinear consolidation with continuous drainage boundaries. Compared with the finite element method (FEM), the PINNs-H achieved comparable accuracy using fewer training data and eliminated the need for space-time discretization. Zhang et al. (2022) developed a physics-informed data-driven approach to automatically recover Terzaghi's consolidation theory from measured data and obtain the true solutions incorporating weak-form PDEs for noise robustness and Monte Carlo dropout for uncertainty estimation. Zhou et al. (2025) proposed a self- adaptive PINN (SA-PINN) method for both forward and inverse problems of nonlinear large strain consolidation, considering the creep characteristics of soft clay. Despite the growing success of PINNs in saturated consolidation problems, however, PINNs have not been applied to unsaturated soil consolidation. First, the governing equations of unsaturated soil consolidation involve coupled partial differential equations representing the simultaneous dissipation of excess pore-air and pore-water pressures, each governed by distinct interactive constant and coefficient of consolidation. This coupling introduces stiffness into the system, leading to imbalance in gradient propagation and training difficulty. Second, the time domain of unsaturated soil consolidation often spans multiple orders of magnitude (s), from rapid air-phase dissipation to slower water-phase drainage over months or years, posing difficulties for conventional PINNs due to their known bias toward fitting long-term dynamics. As a result, training over such long temporal scales often leads to poor convergence, under-resolved late-time predictions, and imbalanced loss propagation unless special strategies are adopted.

This study develops a lagged backward-compatible physics-informed neural network (LBC-PINN) for forward and inverse analysis of one-dimensional unsaturated soil consolidation. The framework incorporates normalization of the spatiotemporal domain, lagged compatibility loss, and transfer learning to improve stability and accuracy across long-time horizons. The forward model predicts coupled pore air-water pressure fields under long-term loading and is validated against FEM simulations. In addition, the paper conducts detailed sensitivity analyses on air-to-water permeability ratios ($k_a/k_w$), time segmentation schemes, neural network structures, and data noise, confirming the robustness and generalizability of the LBC-PINN approach.

## 2. Problem Description

### 2.1 Assumption and Governing Equations

The one-dimensional consolidation model for unsaturated soil proposed by Fredlund and Hasan (1979) is adopted in this study and is illustrated schematically in Fig. 1. A homogeneous, isotropic unsaturated soil layer of thickness $H$ is subjected to a vertical surface load $q$ and extends infinitely in the horizontal direction. Water flow, air flow, and settlement are assumed to occur only in the vertical direction, and air dissolution is ignored in the analysis. The permeability coefficients of water and air, $k_w$ and $k_a$, and the density of water are taken as constant during consolidation. Under these assumptions, the consolidation process involves the coupled dissipation of excess pore-water pressure $u_w$ and excess pore-air pressure $u_a$ under the applied loading q. The corresponding one-dimensional coupled governing equations for $u_a$ and $u_w$ in an isotropic unsaturated soil can be written in the Fredlund-Hasan (1979) form as:

$$\frac{\partial u_a}{\partial t} + C_a \frac{\partial u_w}{\partial t} + c_v^a \frac{\partial^2 u_a}{\partial z^2} = 0 \tag{1}$$

$$\frac{\partial u_w}{\partial t} + C_w \frac{\partial u_a}{\partial t} + c_v^w \frac{\partial^2 u_w}{\partial z^2} = 0 \tag{2}$$

where $C_a = \dfrac{1}{\left[\left(\dfrac{m_1^a}{m_2^a}-1\right) - \dfrac{(1-S_r)n}{m_2^a(u_a^0+u_{atm})}\right]}$ is the interactive constant associated with the air phase; $c_v^a = \dfrac{k_a RT}{gM} \dfrac{1}{\left[m_2^a(u_a^0+u_{atm})\left(\dfrac{m_1^a}{m_2^a}-1\right)-(1-S_r)n\right]}$ is the coefficient of consolidation with respect to the air phase (m²/s); $C_w = \left(\dfrac{m_1^w}{m_2^w} - 1\right)$ is the interactive constant associated with the water phase; $c_v^w = \dfrac{k_w}{m_2^w \gamma_w}$ is the coefficient of consolidation with respect to the water phase (m²/s), $k_a$ and $k_w$ are air and water permeability coefficients (m/s), respectively; $m_1^a$ and $m_2^a$ are coefficients of air volume change with respect to the change in the net normal stress ($\sigma_z - u_a$) and the matric suction ($u_a - u_w$), respectively (/kPa); $m_1^w$ and $m_2^w$ are coefficient of water volume change with respect to the change in the net normal stress ($\sigma_z - u_a$) and the matric suction ($u_a - u_w$), respectively (/kPa); $u_a^0$ and $u_w^0$ are initial pore-air and pore-water pressure (kPa), $u_{atm}$ is the atmospheric pressure (kPa); $R$ is the universal air constant (~8.3 J/mo/K); $T$ is the absolute temperature (K); $M$ is the molecular mass of air saturation during consolidation process; $n$ is the porosity during consolidation process; $S_r$ is the degree of saturation during consolidation process (%); and $\gamma_w$ is water unit weight (~ 9.8 kN/m³). In this study, these material parameters are assumed to be constant with depth and independent of the evolving saturation, so that the governing equations remain linear and directly comparable to existing analytical and numerical solutions.

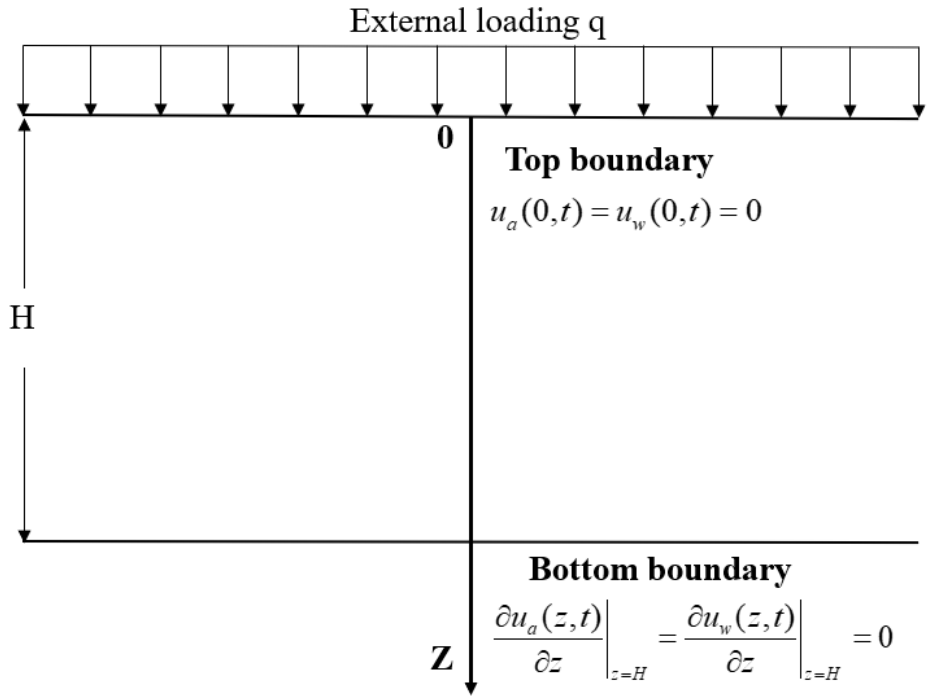

**Fig. 1**. Simplified model for one-dimensional consolidation of the unsaturated soil layer with one-way drainage

## 2.2 Boundary and Initial Conditions

In this study, one-way drainage conditions are evaluated, characterized by a permeable (drained) upper boundary and an impermeable (undrained) bottom boundary, as illustrated in Fig. 1. One-way (single drainage) conditions are imposed for both air and water phases. The top boundary at $z = 0$ is permeable:

$$u_a(0,t) = u_w(0,t) = 0, \quad t > 0; \tag{3}$$

The bottom boundary at $z = H$ is impermeable:

$$\frac{\partial u_a(z,t)}{\partial z}\Big|_{z=H} = \frac{\partial u_w(z,t)}{\partial z}\Big|_{z=H} = 0, \quad t > 0; \tag{4}$$

The initial excess pore-air and pore-water pressures are taken as:

$$u_a(z,0) = u_a^0, u_w(z,0) = u_w^0 \tag{5}$$

## 3. PINN for Unsaturated Soil Consolidation

### 3.1 Standard PINN (STD-PINN)

A physics-informed neural network (PINN) is a machine-learning model that embeds physical laws (usually PDEs) into its loss function, and it learns solutions that satisfy the governing equations even without labeled data (Raissi et al., 2019; Tartakovsky et al., 2020; He and Tartakovsky, 2021; Lu et al., 2021; Mao et al., 2023). By training a neural network to evaluate the approximate solution at arbitrary space-time coordinates and penalizing violations of the governing equations, initial conditions, and boundary conditions through residual loss terms, the standard PINN (STD-PINN) offers a physics-consistent, mesh-free approximation to the solution field (Cai et al., 2021; Toscano et al., 2025). Fig. 2 shows a sketch of the standard PINN framework adopted in this study. Let us consider a general form of a system with m$^{th}$ PDEs defined on a spatiotemporal domain $\Omega \times [0, T]$, where the PDEs describe the evolution of one or more dependent variables $u_i(z,t)$ (e.g., $u_a(z,t)$, $u_w(z,t)$) such that:

$$\mathcal{N}_i[u(z,t); \boldsymbol{\lambda}] = 0, z \in \Omega, t \in (0,T], i = 1,2,\ldots,N \tag{6}$$

Here, $\mathcal{N}_i[\cdot]$ denotes the i$^{th}$ differential operator, which may include mixed partial derivatives in space and time; $\boldsymbol{\lambda} = [\lambda_1, \lambda_2, \ldots]$ are the PDE parameters, where $z$ and $t$ are the space and time coordinates respectively. The corresponding initial and boundary conditions for the PDE system can be expressed as:

$$u_i(z, 0) = u_{i,0}(z), \quad x \in \Omega \tag{7}$$

$$\mathcal{B}_i[u](z, t) = \varphi_i(z, t), \quad z \in \partial\Omega, \quad t \in [0, T] \tag{8}$$

where $u_{i,0}(z)$ represents the initial condition of the PDE, $\partial\Omega$ represents the boundary of the PDE, and $\varphi_i(z, t)$ denotes the corresponding boundary conditions, which can be Dirichlet, Neumann or mixed boundary conditions (Wang et al.,2023).

To approximate the solution fields $u_a$ and $u_w$, a neural network $\hat{u}(\mathbf{z}, t; \boldsymbol{\theta}) = u(\mathbf{z}, t)$ is defined, where $\boldsymbol{\theta}$ denotes the set of trainable parameters (weights and biases) of a fully connected feedforward network (Sharma et al., 2013; Xu et al., 2025). The network typically consists of two input layers, multiple hidden layers with nonlinear activation functions, and two linear output layers. Let the output of the k[th] hidden layer be denoted as $z^{(k)}$, then the network architecture can be expressed as:

$$\begin{cases} \mathbf{z}^0 = (z, t), \\ \mathbf{z}^m = \sigma(\mathbf{W}^m \mathbf{z}^{m-1} + \mathbf{b}^m), m = 1, 2, \dots M \\ \mathbf{z}^{M+1} = \mathbf{W}^{M+1} \mathbf{z}^M + \mathbf{b}^{M+1} \end{cases} \tag{9}$$

where m (m=1,2,...M) refers to the m[th] hidden layer, $\mathbf{z}^m$ is the output of the neurons of the m[th] hidden layer; $\mathbf{W}^m$ and $\mathbf{b}^m$ donate the weight matrix and bias vector of the m[th] layer, $\mathbf{W}^{M+1}$ and $\mathbf{b}^{M+1}$ are the weight matrix and bias vector of the output layer; $\sigma(\cdot)$ is a nonlinear activation function. All the trainable model parameters, i.e., weights and biases, are denoted by $\boldsymbol{\theta}$ in this paper. All required partial derivatives of the network outputs with respect to time and space (e.g., $\partial_t \hat{u}_a$, $\partial_t \hat{u}_w$) are computed through automatic differentiation (Gunes et al., 2015). In PINNs, solving a PDE system is converted into an optimization problem by iteratively updating $\boldsymbol{\theta}$ with the goal of minimizing the loss function $\mathcal{L}_{std}$:

$$\mathcal{L}_{\text{std}}(\boldsymbol{\theta}) = \underbrace{\omega_{\text{IC}} \mathcal{L}_{\text{IC}}(z_k^i, 0) + \omega_{\text{BC}} \mathcal{L}_{\text{BC}}(z_k^b, t_k^b)}_{Physics\ Loss} + \underbrace{\omega_R \mathcal{L}_R(z_k^r, t_k^r)}_{Data\ Loss} \tag{10}$$

where

$$\mathcal{L}_{\mathrm{IC}} = \frac{1}{N_{IC}} \sum_{k=1}^{N_{IC}} \left[ \left(\hat{u}_a(z_k^i, 0) - u_{a,0}(z_k^i, 0)\right)^2 + \left(\hat{u}_w(z_k^i, 0) - u_{w,0}(z_k^i, 0)\right)^2 \right], (z_k^i, 0) \in \Omega \quad (11)$$

$$\mathcal{L}_{\mathrm{BC}} = \frac{1}{N_{BC}} \sum_{k=1}^{N_{BC}} \left[ \left(\hat{u}_a(0, t_k^b)\right)^2 + \left(\hat{u}_w(0, t_k^b)\right)^2 + \left(\frac{\partial \hat{u}_a(H, t_k^b)}{\partial z}\right)^2 + \left(\frac{\partial \hat{u}_w(H, t_k^b)}{\partial z}\right)^2 \right], \quad (12)$$
$$(z_k^b, t_k^b) \in \partial\Omega \times [0, T]$$

$$\mathcal{L}_R = \frac{1}{N_R} \sum_{k=1}^{N_R} \left[ \left( \frac{\partial \hat{u}_a(z_k^r, t_k^r)}{\partial t} + C_a \frac{\partial \hat{u}_w(z_k^r, t_k^r)}{\partial t} + c_v^a \frac{\partial^2 \hat{u}_a(z_k^r, t_k^r)}{\partial z^2} \right)^2 \right.$$
$$\left. + \left( \frac{\partial \hat{u}_w(z_k^r, t_k^r)}{\partial t} + C_w \frac{\partial \hat{u}_a(z_k^r, t_k^r)}{\partial t} + c_v^w \frac{\partial^2 \hat{u}_w(z_k^r, t_k^r)}{\partial z^2} \right)^2 \right], (z_k^r, t_k^r) \in \Omega \times [0, T] \quad (13)$$

where $\mathcal{L}_{\mathrm{IC}}, \mathcal{L}_{\mathrm{BC}}, \mathcal{L}_R$ are initial condition loss, boundary condition loss, and PDE residual loss respectively; $\omega_{\mathrm{IC}}, \omega_{\mathrm{BC}}, \omega_R$ represents the loss weight parameter of $\mathcal{L}_{\mathrm{IC}}, \mathcal{L}_{\mathrm{BC}}, \mathcal{L}_R$, respectively; $N_{\mathrm{IC}}, N_{\mathrm{BC}}, N_R$ are the number of sampling points for each loss term. Physics loss ($\mathcal{L}_{\mathrm{IC}}, \mathcal{L}_{\mathrm{BC}}$) enforces the prescribed initial and boundary conditions, ensuring consistency with problem constraints. The data loss ($\mathcal{L}_R$), defined by the PDE residual, drives the network to satisfy the underlying physical laws throughout the domain (Raissi et al., 2019).

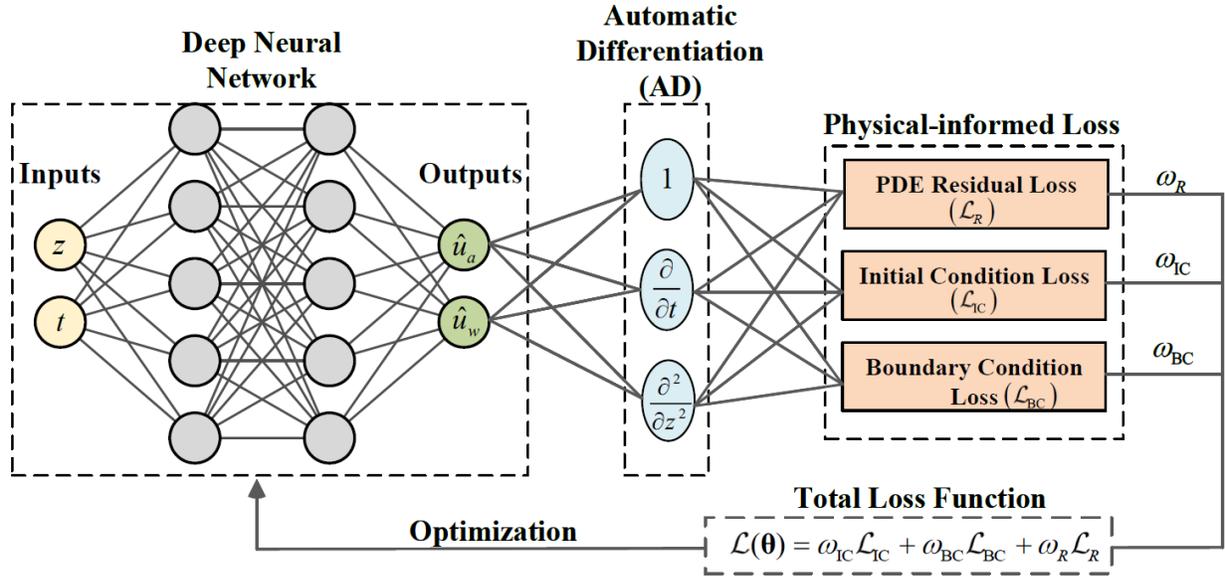

**Fig. 2.** Illustration of the standard physics-informed neural network (STD-PINN).

## 3.2 Lagged Backward-Compatible PINN (LBC-PINN)

Physics-Informed Neural Networks (PINNs) have emerged as a powerful paradigm for solving partial differential equations (PDEs) by embedding physical laws directly into the training of neural networks. However, when applied to problems involving long time scales, such as the consolidation of unsaturated soil, which may span several orders of magnitude in time (e.g., $10^{10}$ s), standard PINNs often exhibit poor convergence, reduced accuracy, or imbalanced resolution across different temporal scales (Chen et al., 2024).

To mitigate these issues, a backward-compatible PINN (BC-PINN) framework is adopted, which is tailored for long-duration consolidation, with several modifications to improve practicality and stability (Mattey and Ghosh, 2022). The time scale is partitioned into segments and trained sequentially with warm starts, as in BC-PINN; however, instead of accumulating residual penalties over all prior segments, a lagged-loss scheme is introduced, which evaluates the objective on the immediately preceding segment when advancing. This backward-looking consistency check, together with explicit interface-continuity enforcement, reduces computational and memory growth while mitigating drift across segments. Firstly, the time domain is divided into $n_{Max}$ contiguous segments (Fig.3):

$$[T_0, T_1], [T_1, T_2], \ldots, [T_{n-1}, T_n], \ldots, [T_{n_{Max}-1}, T_{n_{Max}} = T] \tag{14}$$

where the n$^{th}$ time interval can be present as $\Delta T_n = [T_{n-1}, T_n]$, n =1..., n$_{max}$.

Each segment shares the same spatial domain $\Omega$ but is associated with its own neural network parameters $\boldsymbol{\theta}_n$. For the first subdomain $\Omega \times \Delta T_1$, the standard PINN loss is adopted, which combines the initial condition (IC), boundary condition (BC), and PDE residual terms (as shown in Fig. 3):

$$\mathcal{L}_{\Delta T_1}(\boldsymbol{\theta_1}) = \underbrace{\omega_{IC}\mathcal{L}_{IC}^{(1)}(z_k^i, 0) + \omega_{BC}\mathcal{L}_{BC}^{(1)}(z_k^b, t_k^b)}_{Physics\ Loss} + \underbrace{\omega_R \mathcal{L}_R^{(1)}(z_k^r, t_k^r)}_{Data\ Loss}, \tag{15}$$

$$(z_k^i, 0) \in \Omega, \ (z_k^b, t_k^b) \in \partial\Omega \times [T_0, T_1], (z_k^r, t_k^r) \in \Omega \times [T_0, T_1]$$

where $\mathcal{L}_{IC}^{(1)}$ is the mean-squared error (MSE) of the initial condition at $t = T_0$, $\mathcal{L}_{BC}^{(1)}$ is the MSE of the boundary conditions on $\partial\Omega \times \Delta T_1$, and $\mathcal{L}_R^{(1)}$ is the MSE of the governing PDE residual evaluated at interior collocation points in $\Omega \times \Delta T_1$; $\omega_{IC}, \omega_{BC}$, and $\omega_R$ are constant weights.

For all subsequent sub-domains $\Omega \times$ as $\Delta T_n$, (n = 2,..., $n_{Max}$), a new loss function is proposed, which differs in that it is enforced to satisfy the solution obtained from preceding segment training:

$$\mathcal{L}_{\Delta T_n}(\boldsymbol{\theta_n}) = \underbrace{\omega_{IC}\mathcal{L}_{IC}^{(n)}(z_k^i, T_{n-1}) + \omega_{BC}\mathcal{L}_{BC}^{(n)}(z_k^b, t_k^b)}_{Physics\ Loss} + \underbrace{\omega_R \mathcal{L}_R^{(n)}(z_k^r, t_k^r) + \omega_s \mathcal{L}_S^{(n)}(z_k^s, t_k^s)}_{Data\ Loss},$$

$$(z_k^i, T_{n-1}) \in \Omega \times \{T_{n-1}\}, \ (z_k^b, t_k^b) \in \partial\Omega \times [T_{n-1}, T_n], (z_k^r, t_k^r) \in \Omega \times (T_{n-1}, T_n], \ (z_k^s, t_k^s) \tag{16}$$

$$\in \Omega \times [T_{n-2}, T_{n-1}]$$

Here, $\mathcal{L}_{IC}^{(n)}$, $\mathcal{L}_{BC}^{(n)}$, and $\mathcal{L}_R^{(n)}$ are defined analogously to the first segment but evaluated on $\Omega \times \Delta T_n$. $\omega_s$ controls the strength of the backward-compatibility constraint.

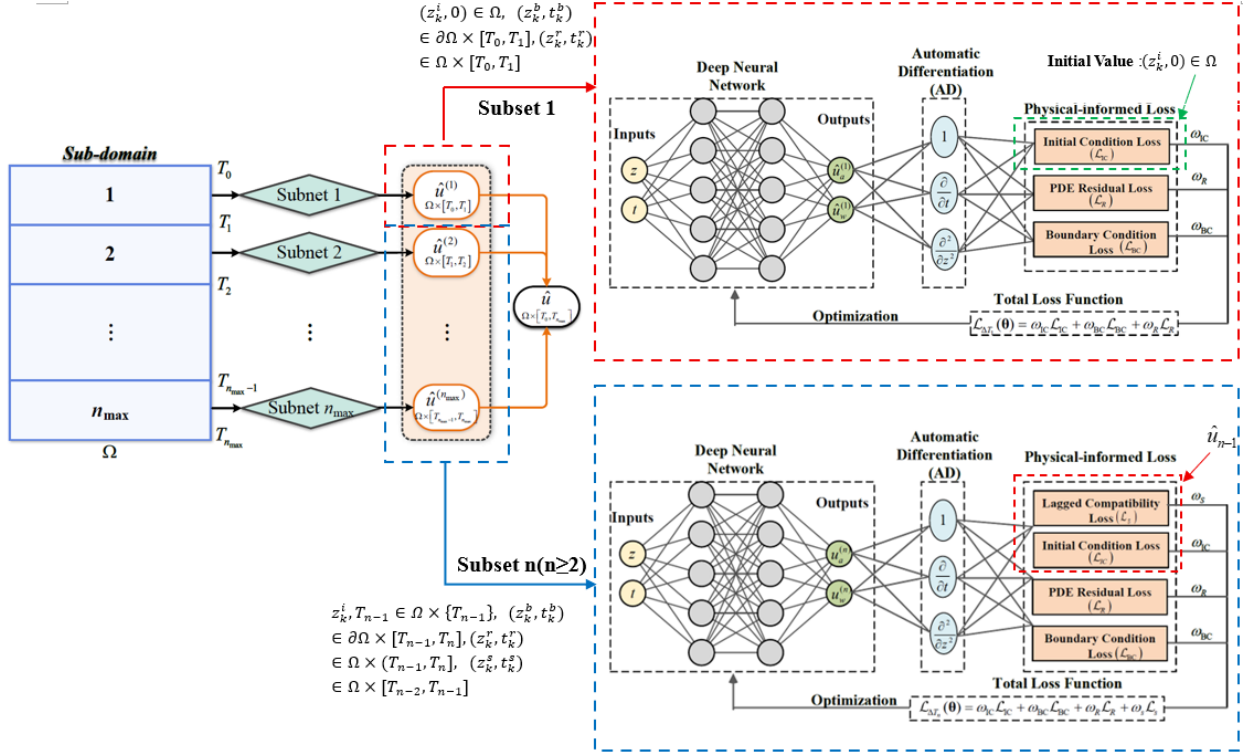

**Fig. 3.** Illustration of the lagged backward-compatible physics-informed neural network (LBC-PINN)

As shown in Fig. 4, the additional lagged compatibility loss term $\mathcal{L}_S^{(n)}$ enforces backward compatibility with the previously trained segment $\Delta T_{n-1} = [T_{n-2}, T_{n-1}]$.

$$\mathcal{L}_S^{(n)}(z_k^s, t_k^s) = \frac{1}{N_S} \sum_{k=1}^{N_S} \left[ \left( \hat{u}_a^{(n)}(z_k^s, t_k^s) - \hat{u}_a^{(n-1)}(z_k^s, t_k^s) \right)^2 \right.$$

$$\left. + \left( \hat{u}_w^{(n)}(z_k^s, t_k^s) - \hat{u}_w^{(n-1)}(z_k^s, t_k^s) \right)^2 \right], (z_k^s, t_k^s) \in \Omega \times [T_{n-2}, T_{n-1}]$$

(17)

where $\hat{u}_a^{(n)}, \hat{u}_w^{(n)}$ and $\hat{u}_a^{(n-1)}, \hat{u}_w^{(n-1)}$ are the pore-air and pore-water pressure predictions of the current and previous segment networks from the preceding time step $\Omega \times [T_{n-2}, T_{n-1}]$. $\mathcal{L}_S$ measures the departure between current predicted solution and the one from the previous training. The introduction of this term ensures backward compatibility.

Finally, the reported LBC-PINN solution is assembled by evaluating the segment-specific models on their respective time intervals and stitching the results to cover the entire spatiotemporal region. Compared with the STD-PINN method, the proposed LBC-PINN additionally employs (i) normalization of the spatiotemporal domain and (ii) segment-wise transfer learning.

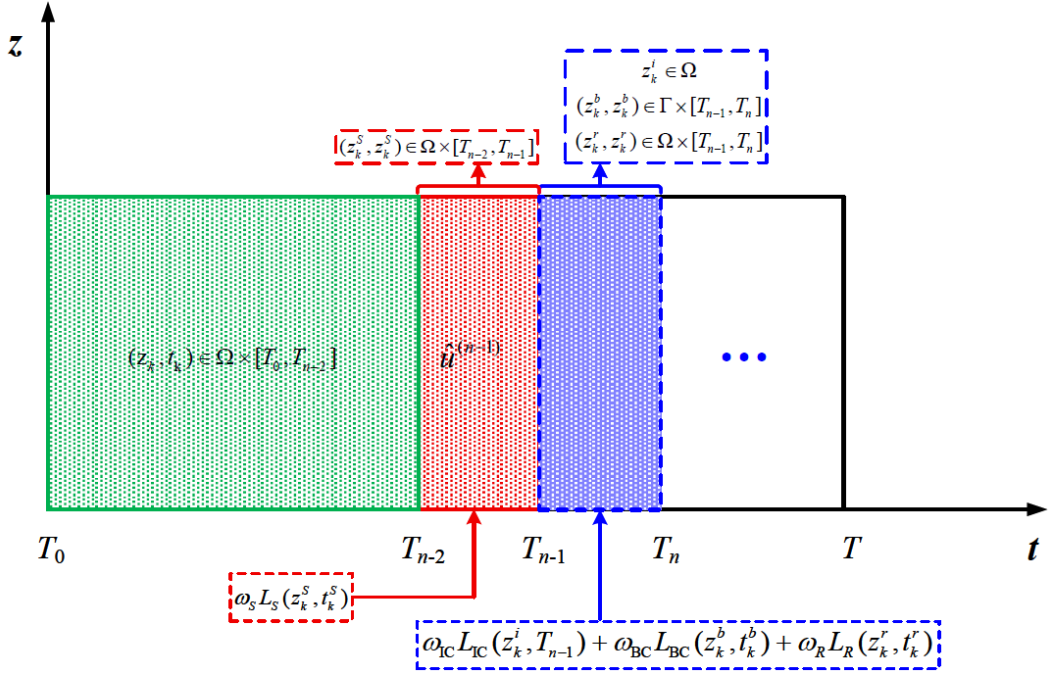

Fig. 3. Illustration of the proposed lagged backward-compatible scheme.

### 3.2.1 Normalization of the spatiotemporal domain

To ensure stable training across multiple time segments, the spatio-temporal domain is normalized into a unit computational domain. Such normalization is a standard preprocessing technique in data-driven deep learning frameworks, and it avoids hidden reweighting effects that arise when raw coordinates span several orders of magnitude (Kumar and Premalatha, 2021; Chen et al.,2024; Xu et al.,2025). In the present study, the depth $z \in [0, H]$ and physical time $t \in [0, T_{max}]$ are first defined in dimensional form. For each time segment $[T_{n-1}, T_n]$, interior collocation points are then generated as follows. Spatial coordinates are sampled uniformly within the layer, $z \in [0, H]$. To achieve balanced resolution over the logarithmic time axis, the base-10 logarithm of physical time is sampled uniformly between $\log_{10} T_{n-1}$ and $\log_{10} T_n$, and mapped back to physical time via $t = 10^{\log_{10} t}$. The resulting pairs $(z, t)$ are finally normalized to the reference domain $\bar{z}, \bar{t} \in [0,1]$ using:

$$\bar{z} = \frac{z - z_{min}}{z_{max} - z_{min}} \tag{18}$$

$$\bar{t} = \frac{t - T_{n-1}}{\Delta T_n} \tag{19}$$

where $z_{min}$ and $z_{max}$ are respectively the minimum and maximum values of z. After this transformation, the governing equations and associated initial and boundary conditions are rewritten in terms of the normalized coordinates $(\bar{z}, \bar{t})$ and enforced on the unit computational domain.

$$\frac{\partial u_a}{\partial \bar{t}} + C_a \frac{\partial u_w}{\partial \bar{t}} + \left(\frac{\Delta T_n}{H^2}\right) c_v^a \frac{\partial^2 u_a}{\partial \bar{z}^2} = 0 \tag{20}$$

$$\frac{\partial u_w}{\partial \bar{t}} + C_w \frac{\partial u_a}{\partial \bar{t}} + \left(\frac{\Delta T_n}{H^2}\right) c_v^w \frac{\partial^2 u_w}{\partial \bar{z}^2} = 0 \tag{21}$$

$$u_a(0, \bar{t}) = u_w(0, \bar{t}) = 0, \quad \bar{t} > 0; \tag{22}$$

$$\frac{\partial u_a(z, \bar{t})}{\partial \bar{z}}|_{\bar{z}=1} = \frac{\partial u_w(z, \bar{t})}{\partial z}|_{\bar{z}=1} = 0, \quad \bar{t} > 0; \tag{23}$$

At the start of segment k ($\bar{t} = 0$, i.e., $t = T_{n-1}$), the initial condition can be written as:

$$u_a(\bar{z}, 0) = u_a(\bar{z}, T_{n-1}), u_w(\bar{z}, 0) = u_w(\bar{z}, T_{n-1}) \tag{24}$$

### 3.2.2 Transfer Learning

Transfer learning is implemented by initializing each LBC-PINN with weights and biases learned from the immediately preceding segment. When moving from segment k-1 to segment k, the network for segment k is initialized with the weights and biases obtained at the end of segment k-1. For the first segment, or when switching to data with a different initial condition, initialization comes from a model trained under that alternative initial condition. Early layers can be frozen briefly for a burn-in period, then unfrozen for full fine-tuning (Goswami et al., 2020; Liu et al.,2023). Nondimensionalization and loss weights remain the same across segments. This setup works with the lagged-loss scheme that evaluates the objective on the immediately preceding segment, improves continuity of $u_a$ and $u_w$ at interfaces, and reduces the number of optimizer steps.

## 4 Implementation

For the forward problem, the architecture of the neural network (STD-PINN and LBC-PINN) has 5 hidden layers with 50 neurons in each layer. For both STD-PINN and LBC-PINN, *tanh* is chosen as the activation function. The advantages of *tanh* activation function are that it is continuous (range $[-1,1]$) and differentiable (Dubey et al., 2022). The neural network has more than 100,000 learning parameters which have been initialized using the "xavier initialization" technique (Glorot and Bengio, 2010). The optimization of the loss function and updating the learning parameters (weights and biases of the neural network) is performed using the L-BFGS optimizer (Kiyani et al., 2025). The number of initial, boundary, and internal sampling points per segment is 2000, 2000, and 10000, respectively. All neural network training was executed in Python 3.12.3 with PyTorch 2.5.0, on an NVIDIA V100 (PCIe; 32 GB HBM2, 5,120 CUDA cores, 640 Tensor Cores). A one-dimensional finite-element model was implemented in COMSOL Multiphysics to generate reference solutions for pore-air and pore-water pressures under the governing equations, boundary and initial conditions described in Section 2. Unless otherwise noted, all STD-PINN and LBC-PINN results are compared against these FEM solutions. For additional verification, the analytical solution of Ho and Fatahi (2014) is evaluated at selected depths and times and overlaid on the FEM and LBC-PINN.

After training PINN, the relative and absolute errors between PINN's predicted solution and the FEM solution are measured. Two evaluation indices, including the mean absolute error (MAE) and the mean relative error (MRE) are used to evaluate the errors between PINN or LBC-PINN and reference results from COMSOL as follows:

$$MAE = \frac{1}{N}\sum_{k=1}^{N}|u_i(z_k,t_k) - \tilde{u}_i(z_k,t_k)| \qquad (25)$$

$$MRE = \frac{1/N \sum_{k=1}^{N}(u_i(z_k,t_k) - \tilde{u}_i(z_k,t_k))}{1/N \sum_{k=1}^{N}(\tilde{u}_i(z_k,t_k))} \tag{26}$$

where $u_i$ is the simulated value predicted by PINN or LBC-PINN (e.g., $u_a, u_w$); $\tilde{u}_i$ is the simulated value of COMSOL solution; $n$ is the sample number. All error metrics are computed with respect to FEM solutions.

## 5. Results and Discussion

This section presents the results with reference to the analytical solution for one-dimensional consolidation of unsaturated soil under one-way drainage conditions, as reported by Ho and Fatahi (2014). The initial thickness H of the soil in the consolidation area is 10 m. The permeability coefficients of both the water ($k_w$) and air ($k_a$) are $10^{-10}$ m/s, the porosity (n) during consolidation process is $n = 0.5$ and the initial degree of saturation ($S_r$) is 80%. The initial pore-air and pore-water pressure is 20 kPa and 40 kPa, respectively. Other relevant soil parameters are listed in Table 1.

**Table 1** Physical property parameters of soil in the forward problem

| n | $S_r$ (%) | $k_a$ (m/s) | $k_w$ (m/s) | $C_a$ | $C_w$ | $c_v^w$ (m²/s) | $c_v^a$ (m²/s) |
|---|---|---|---|---|---|---|---|
| 0.50 | 80 | $10^{-10}$ | $10^{-10}$ | 0.0882 | 0.75 | $6.3 \times 10^{-8}$ | $6.3 \times 10^{-6}$ |

### 5.1. Forward analysis: STD-PINN vs. FEM Results

To evaluate the effectiveness of the LBC-PINN framework, the predictive performance of a standard PINN (STD-PINN) is compared first against the FEM results for 1D consolidation of unsaturated soil. The STD-PINN is trained over the entire time domain $[0, T_{\max}]$ without segmentation. All physics and data losses are evaluated directly on the global domain. Fig. 5 presents STD-PINN predictions trained over progressively extended time horizons $[0, T_{\max}]$, where $T_{\max} \in [10^6, 10^7, 10^8, 10^9, 10^{10}]$s, and compares them with the FEM results for normalized excess pore-air pressure ($u_a/u_{a0}$) and pore-water pressure ($u_w/u_{w0}$).

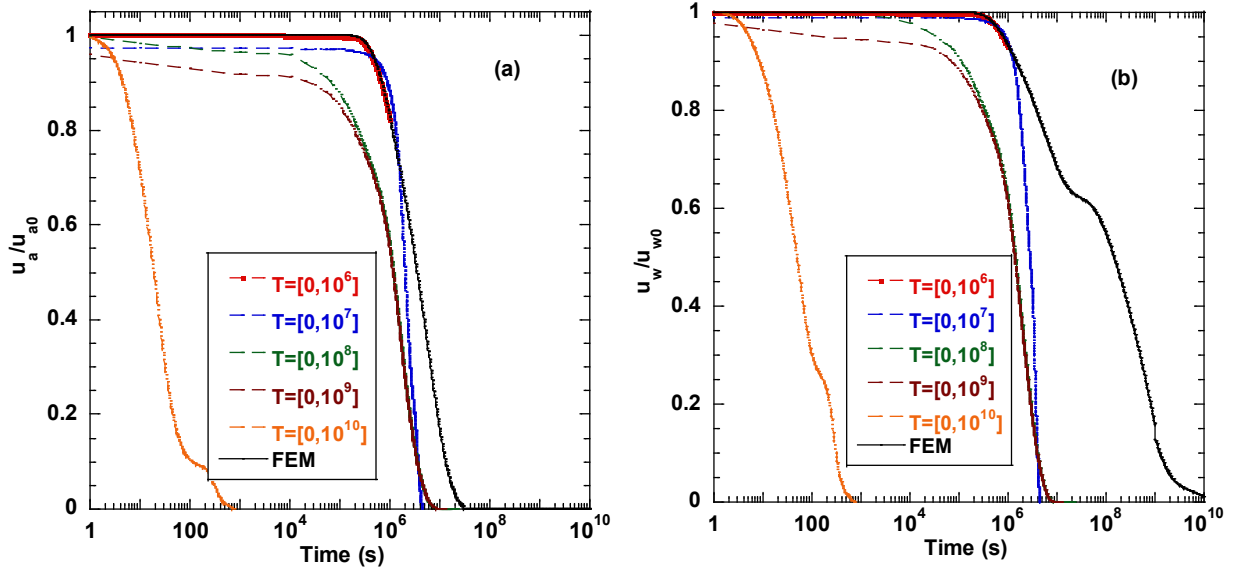

**Fig. 4.** Comparisons between the proposed STD-PINN and FEM for normalized (a) excess pore-air pressure ($u_a/u_{a0}$), (b) excess pore-water pressure ($u_w/u_{w0}$) with different time horizons $[0, T_{\max}]$.

The FEM results exhibit a two-stage dissipation process typical of unsaturated soil consolidation. Initially, the excess pore-air pressure remains nearly constant for an extended period, then dissipates rapidly, followed by a slower dissipation of pore-water pressure. This transition from air-dominated to water-dominated flow spans several orders of magnitude in time, resulting in smooth and physically consistent pressure decay (Fredlund and Hasan, 1979). In contrast, the STD-PINN shows increasing deviations from the reference as $T_{\max}$ grows. At the shortest training horizon of $[0, 10^6]$ s, the STD-PINN captures the overall decreasing trends of both pore-air and pore-water pressures and reproduces the early-time dissipation behavior with reasonable accuracy. STD-PINN model trained on $[0, 10^7]$ s shows temporal lag in both pore-air and pore-water pressure dissipation compared with FEM results. When extended to $[0, 10^8]$ and $[0, 10^9]$ s, the model increasingly underestimates the sharpness and timing of pore-water pressure and pore-water pressure fully dissipation, suggesting that the network struggles to balance early and late time dynamics. By the time the full range $[0, 10^{10}]$ s was used, the STD-PINN showed the greatest deviation from the reference value. The observed degradation in prediction accuracy as $T_{\max}$ increases highlight the limitation of using a single-segment standard PINN for modeling unsaturated soil consolidation.

The deep networks exhibit a well-documented spectral bias toward low-frequency/slowly varying components, which makes them under-resolve the sharp early-time dissipation and instead produce a delayed, overly steep drop when the training window spans many decades of time (Rahaman et al.,2019). Also, PINNs show marked discrepancies in convergence rates and ill-conditioning across loss terms, so the optimizer gravitates to "easy" late-time portions of the domain and neglects early transients, which worsens as the global horizon [0, $T_{max}$] grows (Wang et al., 2022).

## 5.2. Forward analysis: LBC-PINN vs. FEM Results

Fig. 6 presents the evolution of total, data, and physics losses during the LBC-PINN training process for four representative time segments (total time segment N=5). The consistent drop in both loss components (e.g., data loss, physics loss) across segments confirms the stability and adaptability of the LBC-PINN framework when solving long-term unsaturated soil consolidation problems. The relative contributions of data and physics losses evolve notably as the time window shifts from early to late stages. In the earliest segment ($0 \leq t \leq 10^2$ s; Fig. 6a), the physics loss dominates and closely tracks the total loss. This behavior reflects the fact that, at early times, the dissipation of pore-air and pore-water pressure is primarily constrained by the initial and boundary conditions, as minimal pore-air, pore-water pressure dissipation has occurred and the PDE residual remains small. As time advances (Figs. 6b–6c), the data loss begins to contribute more significantly to the total loss. In the final segment ($10^8 \leq t \leq 10^{10}$ s; Fig. 6d), the data loss becomes the dominant component, aligning closely with the total loss, while the physics loss remains lower and more stable. This shift indicates that, as the system progresses into the water-dominated dissipation phase, the solution is increasingly governed by the internal dynamics described by the PDEs, and less by the boundary and initial constraints.

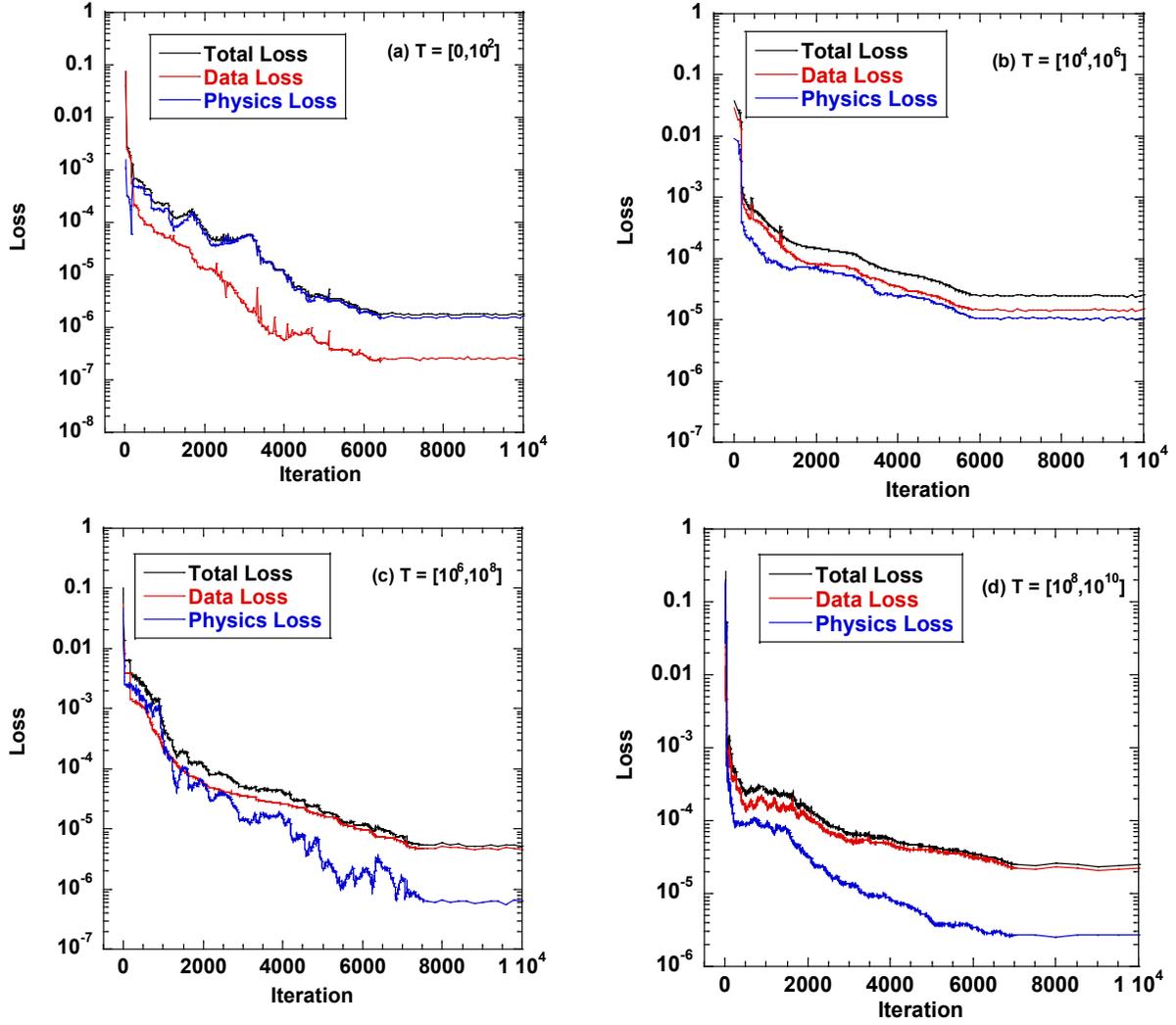

**Fig. 5.** Evolution of Loss with iterations of the LBC-PINN under (a) $T = [0,10^2]$, (b) $T = [10^4,10^6]$, (c) $T = [10^6,10^8]$, (d) $T = [10^8,10^{10}]$.

Fig. 7 compares LBC-PINN predictions with FEM results at depth z = 2.5, 5.0, and 7.5 m for normalized excess pore-air and pore-water pressures. Across all depths, the LBC-PINN closely follows the FEM benchmark throughout the entire time domain. It accurately captures the depth-dependent delay in the onset of dissipation for both pore-air and pore-water pressures and reproduces the multi-stage S-shaped decay of $u_w/u_{w0}$, including the characteristic shoulder near $10^7 \sim 10^8$s. The square markers extracted from Ho and Fatahi (2014) align well with both the FEM and LBC-PINN results across all three depths.

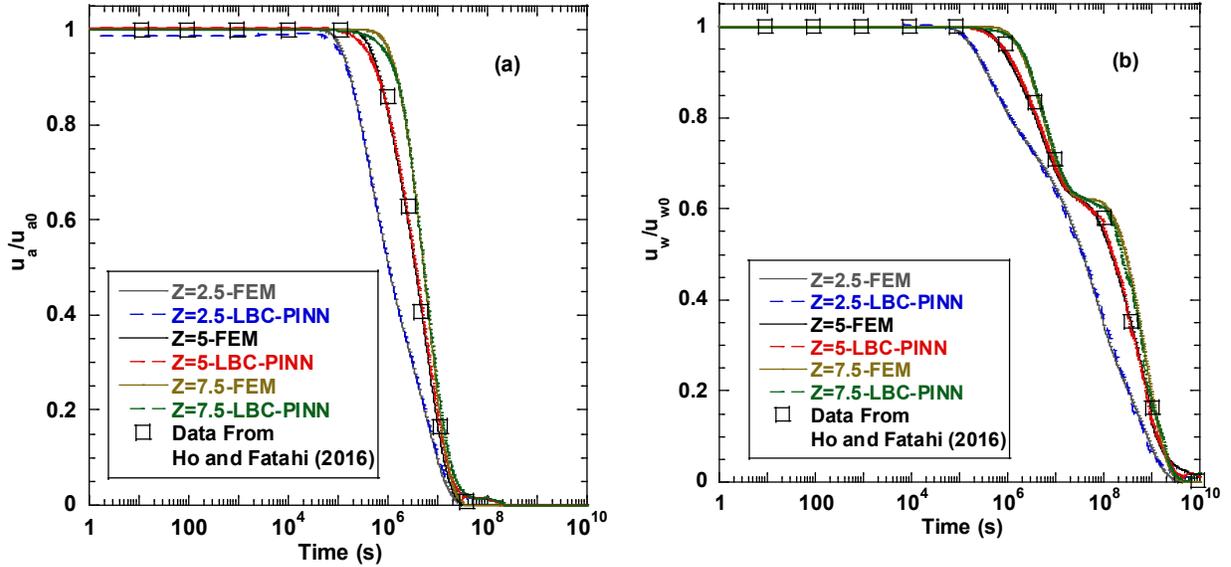

**Fig. 6.** Comparison of the (a) ($u_a/u_{a0}$), (b) ($u_w/u_{w0}$) between LBC-PINN and FEM results for selected depth.

Similarly, Fig. 8 demonstrates the dissipation of pore-air and pore-water pressures with depth at different time factors $T_v$ ($T_v = k_w t / m_s^1 \gamma_w H^2$). The LBC-PINN results (dashed lines) show strong agreement with the finite element (FEM) solution (solid lines) across all time snapshots. Fig. 9 compares the spatio-temporal fields of normalized pore-air and pore-water pressures predicted by the LBC-PINN and FEM. The LBC-PINN show excellent agreement with FEM results across the domain. Absolute error maps indicate low errors concentrated near the early-time boundary regions, with no significant accumulation at later times. For pore-air pressure, larger errors are observed at intermediate times and mid-depths, while for pore-water pressure, noticeable deviations occur at later times in deeper zones. Peak errors remain below 0.05, confirming the accuracy of the LBC-PINN in capturing both phases over space and time.

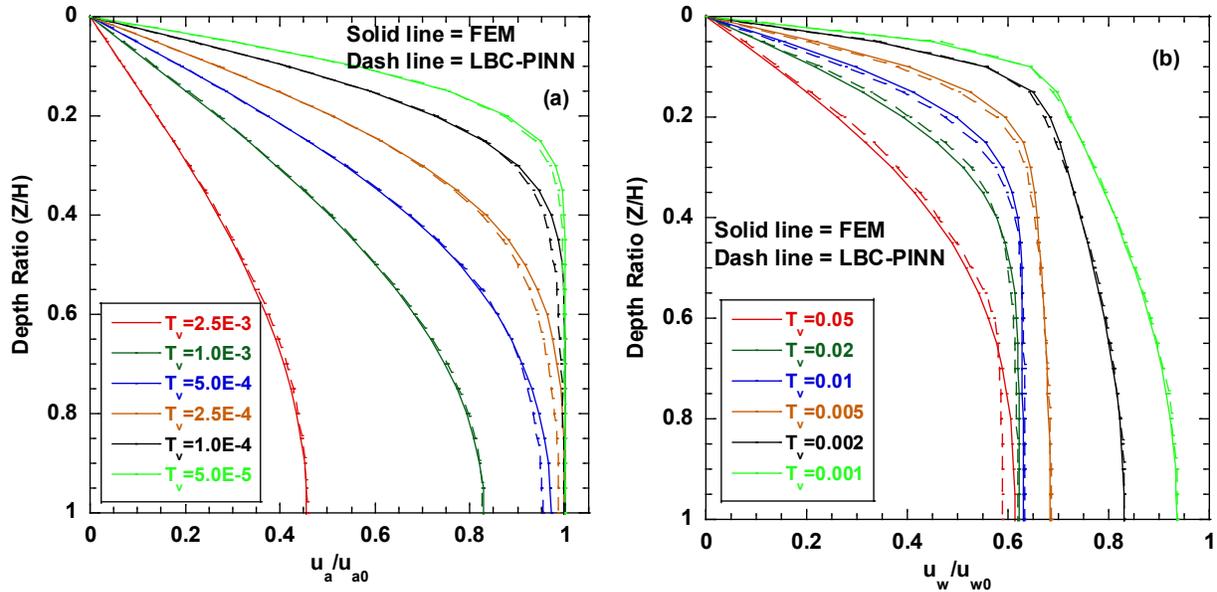

**Fig. 7.** Comparison of the (a) ($u_a/u_{a0}$), (b) ($u_w/u_{w0}$) between LBC-PINN and FEM results for selected $T_v$.

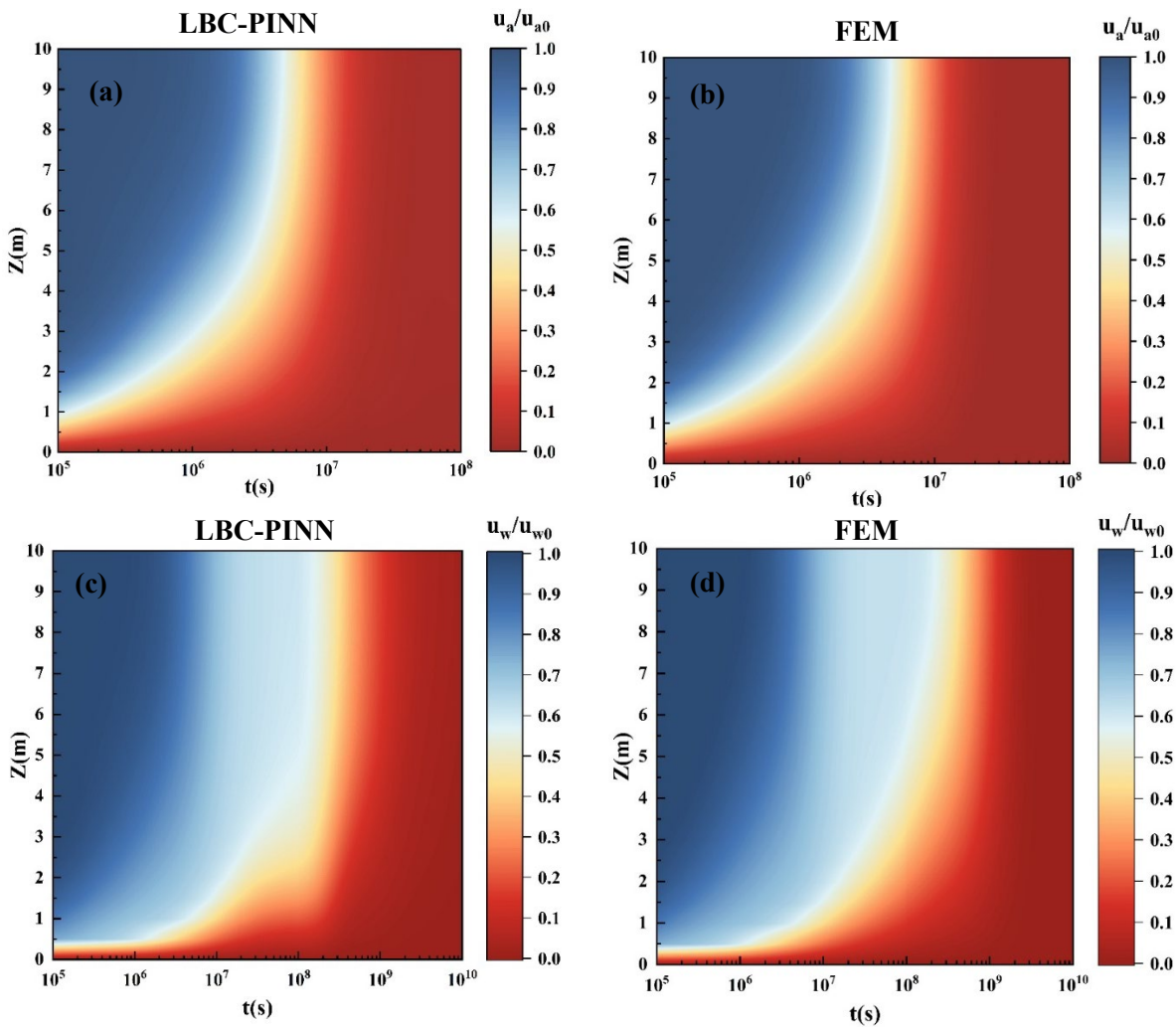

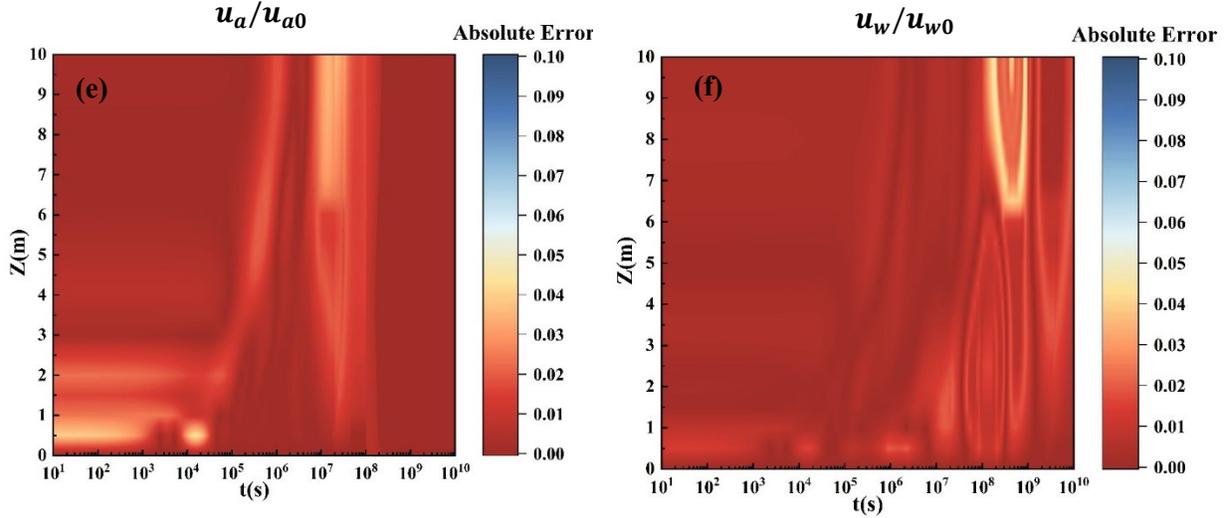

**Fig. 8.** Spatiotemporal contours of (a) $(u_a/u_{a0})$ by LBC-PINN, (b) $(u_a/u_{a0})$ by FEM, (c) $(u_w/u_{w0})$ by LBC-PINN, (d) $(u_w/u_{w0})$ by FEM. The absolute error between the LBC-PINN and FEM solutions of (e) $(u_a/u_{a0})$, (f) $(u_w/u_{w0})$.

## 5.3 Sensitivity Analysis

### 5.3.1 Influence of Time Segmentation

To evaluate the influence of time segmentation on model accuracy, the segmented LBC-PINN is trained with increasing numbers of log-scaled time partitions ($N$ = 2 to 6) over a fixed simulation horizon [0, $10^{10}$]. As shown in Table 2 and Fig.10, the case with $N$ = 2 exhibits the poorest performance, with a combined mean absolute error (MAE) of 0.233, mean relative error (MRE) of 0.321, and a peak absolute error exceeding 1.0. The coefficient of determination ($R^2 = 0.116$) indicates that the model fails to capture the sharp gradients and multiscale behavior inherent in unsaturated soil consolidation problem. Increasing to three segments (N = 3) yields a substantial reduction in MAE (0.028) and improved $R^2$(0.961), although the MRE remains high (0.123), primarily due to relative error amplification in both pore-air pressure and pore-water pressure $u_w$ at later times where true values (shown in Fig.11). Once the number of time segments reaches four ($N = 4$), the model achieves high accuracy across both pore-air and pore-water pressure phases, with a combined MAE under 0.011, MRE around 0.059, and R² surpassing 0.9995 for both $u_a$ and $u_w$. The results $N = 5$, and $N = 6$ are nearly identical, with all key metrics showing minimal variation. This plateau suggests that five segments are sufficient to resolve the temporal complexity of the problem. The evolution of relative error over time and segmentation is shown Fig. 11. For $u_a$, relative error is primarily concentrated in the intermediate time range ($10^6$~$10^8$ s), where rapid dissipation and coupling with $u_w$ induce steep gradients. In contrast, $u_w$ shows a gradual rise in relative error beyond $10^6$ s.

The improvement with increasing $N$ is consistent with known limitations of PINNs on long-horizon, stiff PDEs. Neural networks tend to learn low-frequency components more easily, known as spectral bias (Rahaman et al., 2019), which leads to underfitting of sharp early-time dynamics when trained over wide temporal domains. Additionally, coarse segmentations suffer from gradient imbalance, where early-time physics losses are overshadowed by smoother long-time behavior, degrading convergence and accuracy (Wang et al., 2021). By dividing the time domain into more segments, each subproblem spans a narrower temporal range, improving approximation quality, optimizer conditioning, and sampling effectiveness. This approach aligns with prior work on domain decomposition and sequential learning, including XPINNs (Jagtap et al., 2020) and bc-PINNs (Mattey and Ghosh, 2022), which advocate for solving multi-scale PDEs through staged, backward-compatible training. Overall, segmenting the time domain into five or more subintervals transforms a stiff, long-term learning task into a series of well-posed problems, with $N = 5$ emerging as an optimal balance between accuracy and computational efficiency.

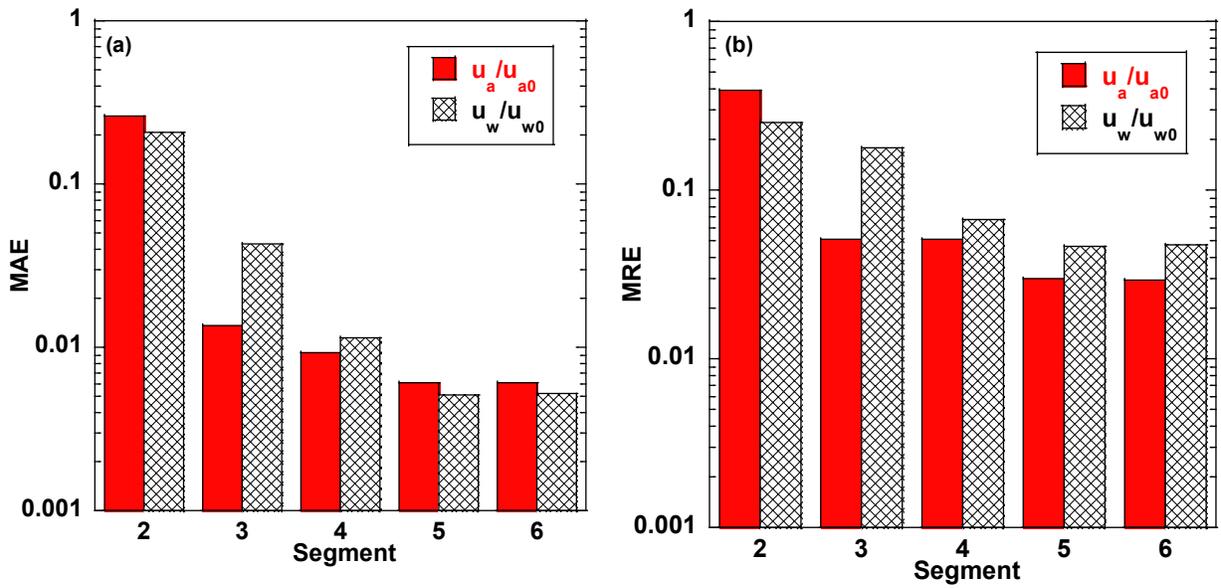

**Fig. 9.** Accuracy vs. time segmentation level ($N$): (a) MAE and (b) MRE for $(u_a/u_{a0})$ and $(u_w/u_{w0})$.

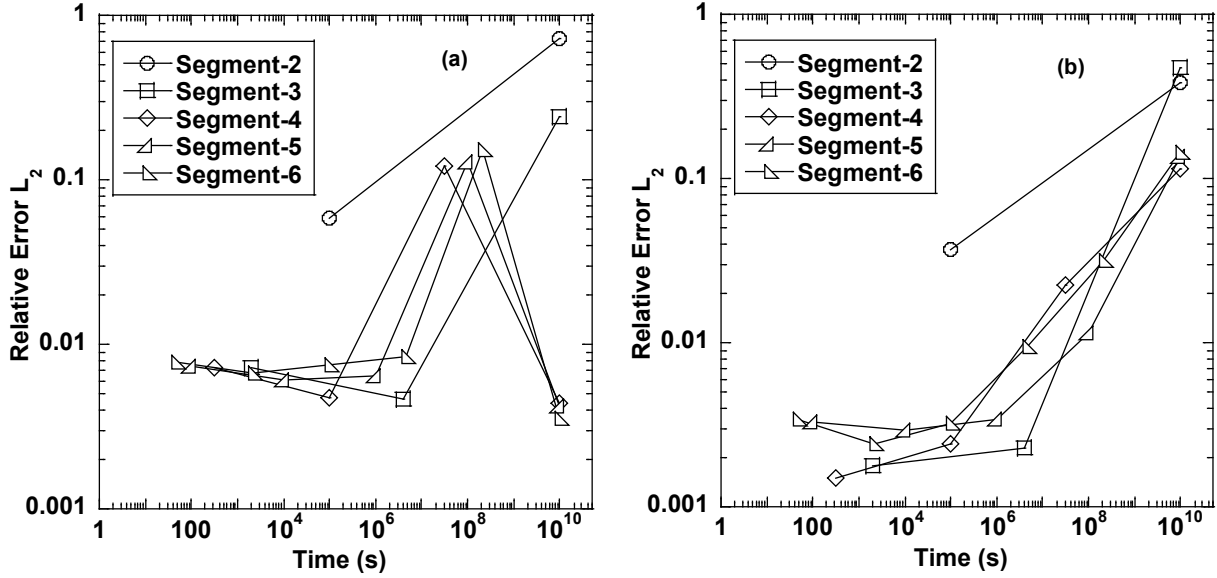

**Fig. 10.** Relative Error across different training segments over Time: (a) ($u_a/u_{a0}$), (b) ($u_w/u_{w0}$).

**Table 2** Comparisons between the FEM and LBC-PINN with time segmentation level

| Time Segment (N) | Variable | MAE | MRE | Max Abs. Error | R² |
|---|---|---|---|---|---|
| 2 | $u_a$ | 0.2598 | 0.3880 | >1.0 | 0.1271 |
| | $u_w$ | 0.2071 | 0.2529 | >1.0 | 0.0483 |
| | Combined ($u_a+u_w$) | 0.2334 | 0.3205 | >1.0 | 0.1157 |
| 3 | $u_a$ | 0.0135 | 0.0516 | 0.2796 | 0.9921 |
| | $u_w$ | 0.0429 | 0.1779 | 0.5352 | 0.9098 |
| | Combined ($u_a+u_w$) | 0.0282 | 0.1226 | 0.5352 | 0.9606 |
| 4 | $u_a$ | 0.0093 | 0.0514 | 0.1041 | 0.9995 |
| | $u_w$ | 0.0114 | 0.0668 | 0.0493 | 0.9995 |
| | Combined ($u_a+u_w$) | 0.0114 | 0.0591 | 0.0767 | 0.9995 |
| 5 | $u_a$ | 0.0061 | 0.0301 | 0.0421 | 0.9995 |
| | $u_w$ | 0.0051 | 0.0469 | 0.0485 | 0.9995 |
| | Combined ($u_a+u_w$) | 0.0056 | 0.0385 | 0.0453 | 0.9995 |
| 6 | $u_a$ | 0.0061 | 0.0291 | 0.0432 | 0.9995 |
| | $u_w$ | 0.0052 | 0.0477 | 0.0488 | 0.9996 |
| | Combined ($u_a+u_w$) | 0.0054 | 0.0384 | 0.0469 | 0.9995 |

**Notes:** MAE = Mean Absolute Error; MRE = Mean Relative Error; R² = coefficient of determination. Combined = metrics computed on concatenated $u_a$ and $u_w$ series.

### 5.3.2 Impact of Air-to-water Permeability Ratios

The pore-air and pore-water pressures predicted by the proposed LBC-PINN and FEM across air-to-water permeability ratios of $k_a/k_w$ were shown in Fig.12. The results showed the air pressure decay curves for different permeability ratios form parallel trends, where higher $k_a/k_w$ values result in earlier and steeper dissipation trend. For instance, at $k_a/k_w=10.0$, pore-air pressure dissipates rapidly and is nearly fully released by approximately $5 \times 10^6$ seconds, whereas the pore-air pressure fully released by $5 \times 10^7$ seconds with $k_a/k_w=1.0$. As shown in Fig. 12b, pore-water pressure dissipation under one-way drainage follows a two-stage process. Initially, pore-water pressure decreases rapidly, especially at higher $k_a/k_w$, due to simultaneous air outflow, forming an "upper S"-shaped curve. Once pore-air pressure fully dissipates, all pore-water pressure curves converge and gradually decline, resembling classical saturated-soil consolidation. This second stage extends to approximately $10^{10}$ seconds. The LBC-PINN closely matches the data across all cases, accurately capturing both the early-stage dissipation and the long-term consolidation behavior over time scales up to $10^{10}$ seconds. Minor phase lag is observed in the low-permeability case ($k_a/k_w = 0.1$), but the overall trends and magnitudes are well preserved.

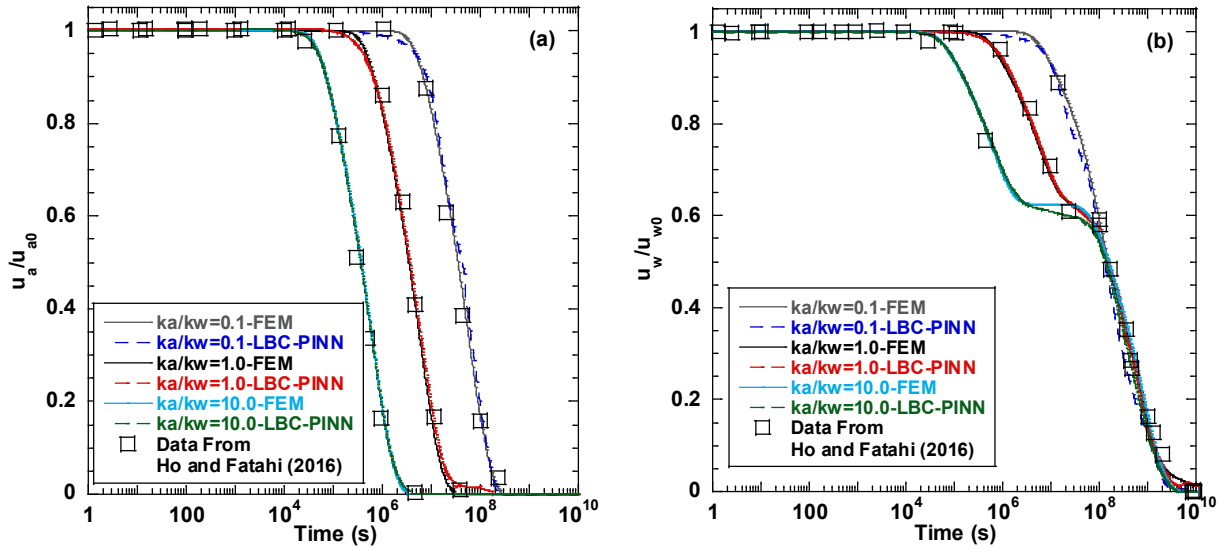

**Fig. 11.** Comparison of the (a) ($u_a/u_{a0}$), (b) ($u_w/u_{w0}$) between LBC-PINN and FEM results across air-to-water permeability ratios of $k_a/k_w$.

Fig. 13a and 13b show the variation of mean absolute error (MAE) and mean relative error (MRE) in the predicted pore-air and pore-water pressures as a function of $k_a/k_w$, ranging from 0.001 to 1000. Overall, the LBC-PINN achieves low error levels across all cases, with both MAE and MRE values consistently below 0.2. However, noticeably higher errors are observed when $k_a/k_w < 1$, particularly at $k_a/k_w = 0.1$, where the model exhibits the largest MAE and MRE for both $u_a$ and $u_w$. The reduced accuracy of the LBC-PINN at $k_a/k_w < 1$, especially around 0.1, is due to the stronger coupling between air and water pressures. Low air permeability slows down air pressure dissipation, causing both $u_a$ and $u_w$ to evolve simultaneously over extended periods. The PINN network are particularly sensitive to such multiscale, coupled dynamics due to limitations in learning sharp transients and interacting residuals under stiff conditions (Wang et al., 2021). In contrast, when $k_a/k_w \geq 1$, the pore-air pressure dissipates more quickly and decouples from the system early on. As a result, excess pore-water pressure dissipation follows a simpler, smoother consolidation path, allowing the network to achieve better accuracy and faster convergence.

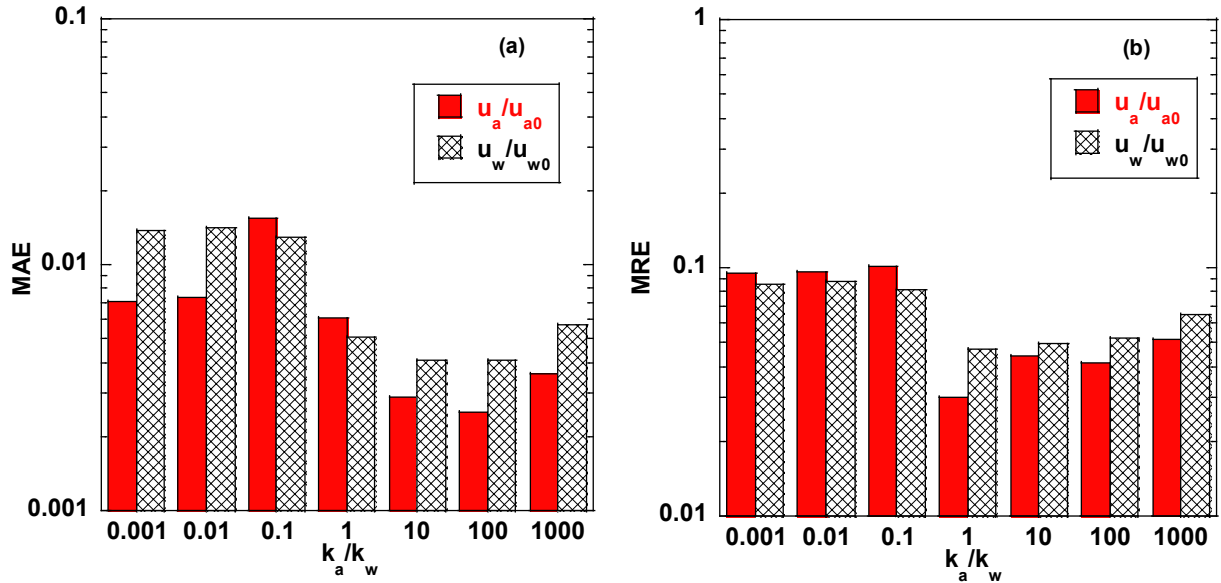

**Fig. 12.** Accuracy vs. air-to-water permeability ratios of $k_a/k_w$: (a) MAE and (b) MRE for $(u_a/u_{a0})$ and $(u_w/u_{w0})$.

Fig.14 illustrates normalized settlement ($s^*$) with time predicted by the LBC-PINN and FEM solutions under one-way drainage conditions. The results show excellent agreement between the LBC-PINN and FEM across all cases, demonstrating the model's ability to accurately capture the overall consolidation behavior of unsaturated soil over a wide temporal range (up to $10^{10}$s). As expected, increasing $k_a/k_w$ shifts the settlement curve toward earlier times, indicating a faster consolidation response in this setup.

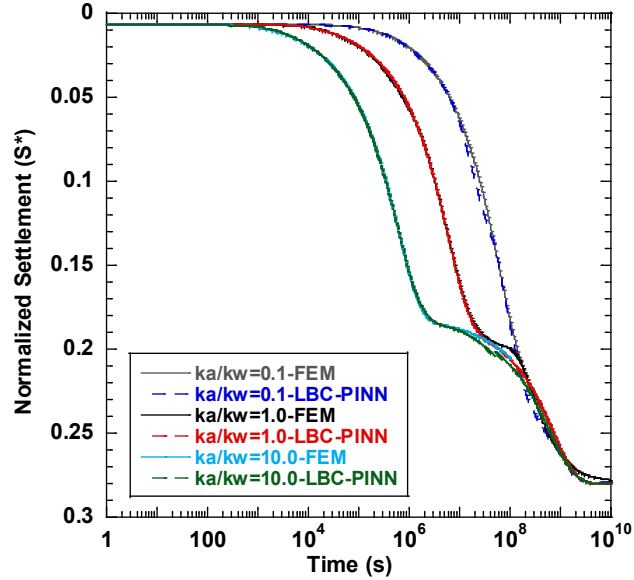

**Fig. 13.** Comparison of the normalized settlement (S*) between LBC-PINN and FEM results across air-to-water permeability ratios of $k_a/k_w$.

**Table 3** Comparisons between the FEM and LBC-PINN with different air-to-water permeability ratios of $k_a/k_w$.

| $k_a/k_w$. | Variable | MAE | MRE | Max Abs. Error | R² |
|---|---|---|---|---|---|
| 0.001 | $u_a$ | 0.0071 | 0.0947 | 0.0541 | 0.996 |
| | $u_w$ | 0.0138 | 0.0854 | 0.0523 | 0.996 |
| | Combined ($u_a+u_w$) | 0.0104 | 0.0901 | 0.0532 | 0.996 |
| 0.01 | $u_a$ | 0.0074 | 0.0957 | 0.0514 | 0.9960 |
| | $u_w$ | 0.0142 | 0.0877 | 0.0502 | 0.9959 |
| | Combined ($u_a+u_w$) | 0.0108 | 0.0917 | 0.0508 | 0.9960 |
| 0.1 | $u_a$ | 0.0155 | 0.1019 | 0.0463 | 0.9972 |
| | $u_w$ | 0.0129 | 0.0811 | 0.0484 | 0.9964 |
| | Combined ($u_a+u_w$) | 0.0142 | 0.1019 | 0.0484 | 0.9969 |
| 1.0 | $u_a$ | 0.0061 | 0.0301 | 0.0421 | 0.9995 |
| | $u_w$ | 0.0051 | 0.0469 | 0.0485 | 0.9995 |

| $k_a/k_w$ | Variable | MAE | MRE | Max Abs. Error | R² |
|---|---|---|---|---|---|
| | Combined ($u_a+u_w$) | 0.0056 | 0.0385 | 0.0453 | 0.9995 |
| 10.0 | $u_a$ | 0.0029 | 0.0441 | 0.0501 | 0.9997 |
| | $u_w$ | 0.0041 | 0.0498 | 0.0787 | 0.9984 |
| | Combined ($u_a+u_w$) | 0.0035 | 0.0469 | 0.0419 | 0.9987 |
| 100 | $u_a$ | 0.0025 | 0.0414 | 0.0542 | 0.9950 |
| | $u_w$ | 0.0041 | 0.0521 | 0.0414 | 0.9980 |
| | Combined ($u_a+u_w$) | 0.0033 | 0.0467 | 0.0478 | 0.9965 |
| 1000 | $u_a$ | 0.0036 | 0.0515 | 0.0485 | 0.9962 |
| | $u_w$ | 0.0057 | 0.0644 | 0.0687 | 0.9941 |
| | Combined ($u_a+u_w$) | 0.0046 | 0.0629 | 0.0486 | 0.9951 |

**Notes:** MAE = Mean Absolute Error; MRE = Mean Relative Error; R² = coefficient of determination. Combined = metrics computed on concatenated $u_a$ and $u_w$ series.

### 5.3.3 Effect of structure of neural network

In LBC-PINN, the number of hidden layers (depth) and neurons per layer (width), has a significant influence on prediction accuracy. To evaluate the sensitivity of the model to these hyperparameters, a range of neural network configurations was tested, with hidden layers set to 1, 3, 5, 7, and 10, and neurons per layer set to 10, 30, 50, 70, and 100. As shown in Fig. 15a, the model with 5 hidden layers and 50 neurons per layer achieved the lowest relative $L_2$ error, indicating the best performance among all configurations. In general, increasing the network size led to improved accuracy; however, once the network exceeded a moderate size (e.g., more than 5 layers or 50 neurons), the error values plateaued and remained within the same order of magnitude. The networks with only a single hidden layer consistently exhibited higher errors, highlighting the necessity of sufficient depth for effective learning in this application.

### 5.3.4 Effect of training data

To investigate the influence of collocation point density on the accuracy of the LBC-PINN framework, a sensitivity study was conducted by varying the number of interior (residual) and boundary collocation points per segment. Fig.15b presents a heatmap of the relative $L_2$ error under different combinations of residual points (ranging from 1,000 to 20,000) and boundary points (ranging from 50 to 2,000).

The results indicate that increasing the number of residual points generally improves accuracy, particularly when moving from 1,000 to 10,000 points. Notably, the lowest $L_2$ error (0.0385) is achieved with 1,000 boundary points and 20,000 residual points. However, increasing boundary points beyond 1,000 offers diminishing returns and, in some cases (e.g., 2000 points), introduces instability, likely due to overfitting. Conversely, models with too few residual points (e.g., 1,000) exhibit substantially higher errors regardless of boundary point count, confirming the critical role of sufficient interior constraint enforcement in physics-informed learning.

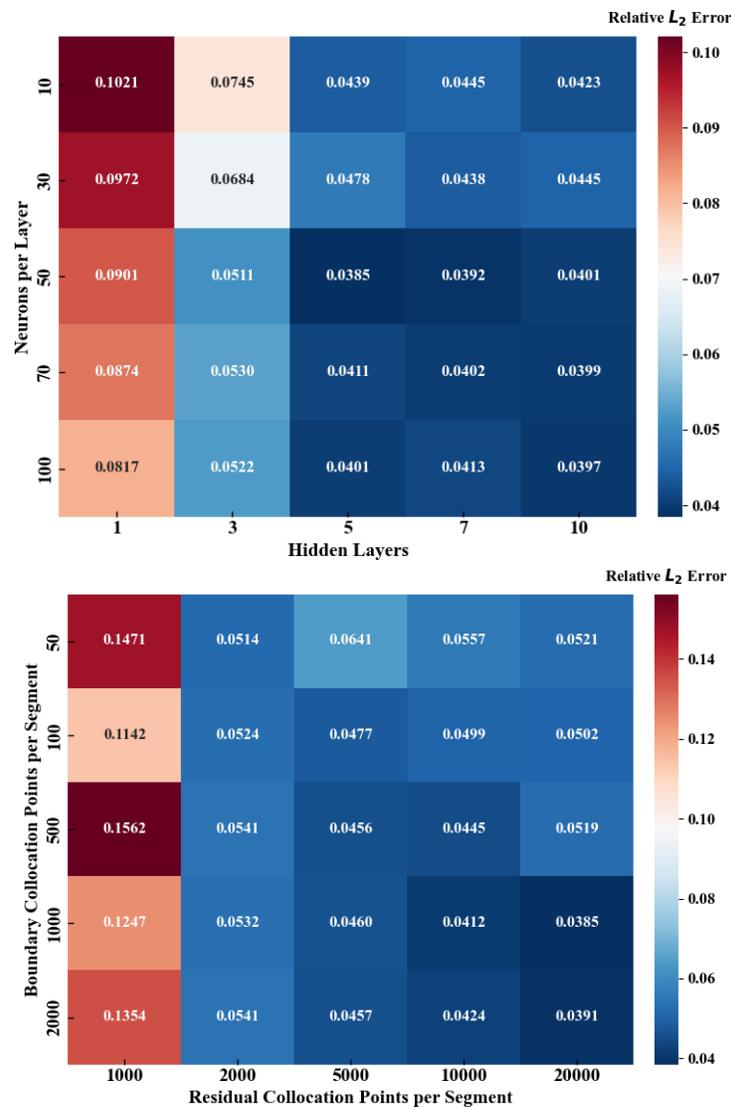

**Fig. 14.** Average relative errors of the learning results of 1D unsaturated soil consolidation under different (a) Neural network structures and (b) Training data amount.

## 5.4 Simplified Time Segmentation Strategy

As shown in Fig. 7, at the early-time, excess pore-air pressure remains nearly constant, followed by a sharp drop around $10^6$ s as the dissipation initiates, which can be computationally expensive to resolve accurately, especially if these dynamics occur before dissipation of pore-water pressure. To reduce computational cost and improve training efficiency, a simplified strategy was proposed based on the characteristic time for pore-air pressure full dissipation, expressed as $t_s = H^2/c_v^a$, where $H$ is the drainage path length and $c_v^a$ is the air-phase consolidation coefficient. This time ($t_s$) is assumed to represent the point at which pore-air pressure has fully dissipated. To ensure consistency across cases, the calculated time was rounded down to the nearest order of magnitude and used as the starting point for time segmentation. The remainder of the time domain was then partitioned following the same approach adopted in the baseline LBC-PINN.

Fig. 16 presents the absolute error distributions of predicted pore-air and pore-water pressures ($u_a$ and $u_w$) using the simplified time segmentation strategy. For $k_a/k_w = 0.1$, larger $u_a$ errors emerge near the bottom boundary with time ranged from $10^5$ to $10^7$s, attributed to delayed pore-air dissipation and strong air–water pressure coupling. The corresponding $u_w$ errors also concentrate near dissipation fronts, especially at later stages, indicating the difficulty in resolving coupled transient responses. At $k_a/k_w = 1.0$, both $u_a$ and $u_w$ errors are moderate and more uniformly distributed, reflecting simultaneous dissipation. When $k_a/k_w = 10.0$, $u_a$ errors diminish rapidly due to faster air flow, while $u_w$ errors persist near the bottom and late times, where pore-water controls the dominant dissipation.

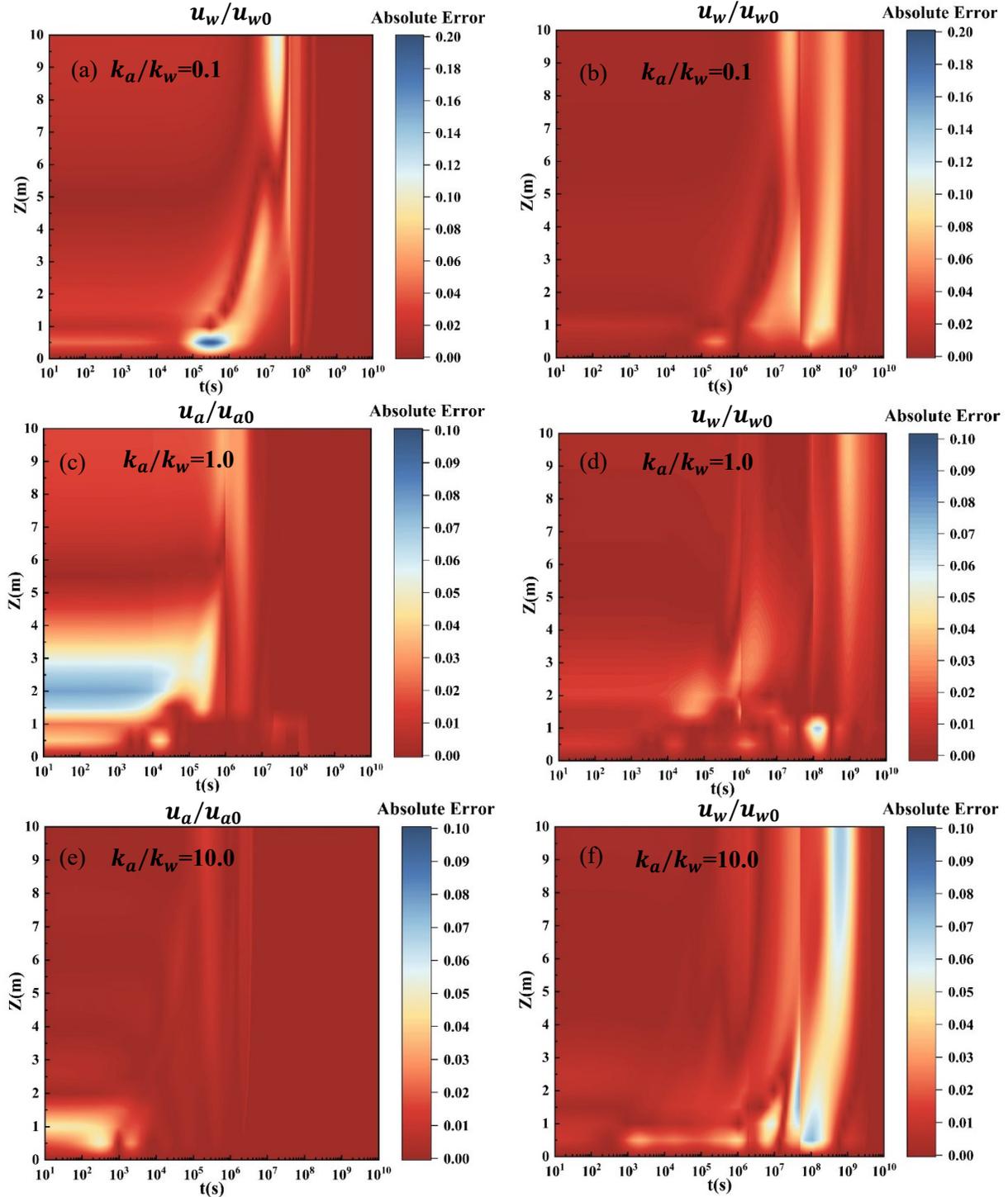

**Fig. 15.** The absolute error for the LBC-PINN and FEM solutions with simplified time segment partition strategy of (a) ($u_a/u_{a0}$) with $k_a/k_w$= 0.1, (b) ($u_w/u_{w0}$) with $k_a/k_w$= 0.1, (c) ($u_a/u_{a0}$) with $k_a/k_w$= 1.0, (d) ($u_w/u_{w0}$) with $k_a/k_w$= 1.0, (e) ($u_a/u_{a0}$) with $k_a/k_w$= 10.0, (f) ($u_w/u_{w0}$) with $k_a/k_w$= 10.0,

Fig. 17 quantitatively confirms the trends observed in the absolute error maps. When $k_a/k_w \leq 0.1$, the combined mean absolute error (MAE) and mean relative error (MRE) for $u_a$ and $u_w$ are noticeably higher, reaching a peak MAE of 0.0169 and MRE of 0.1764 at $k_a/k_w = 0.01$. This elevated error is attributed to the strong air–water pressure coupling in this regime, where both pressures evolve simultaneously and interact nonlinearly, making the solution space more complex and harder for the PINN to approximate. As $k_a/k_w$ increases beyond 1, both MAE and MRE for both $u_a$ and sharply decrease. For instance, the MAE for $u_a$ drops from 0.0174 at $k_a/k_w = 0.001$ to 0.0032 at $k_a/k_w = 10$. The MRE for $u_w$ drops from 0.1824 at $k_a/k_w = 0.01$, to 0.0787 at $k_a/k_w = 10$.

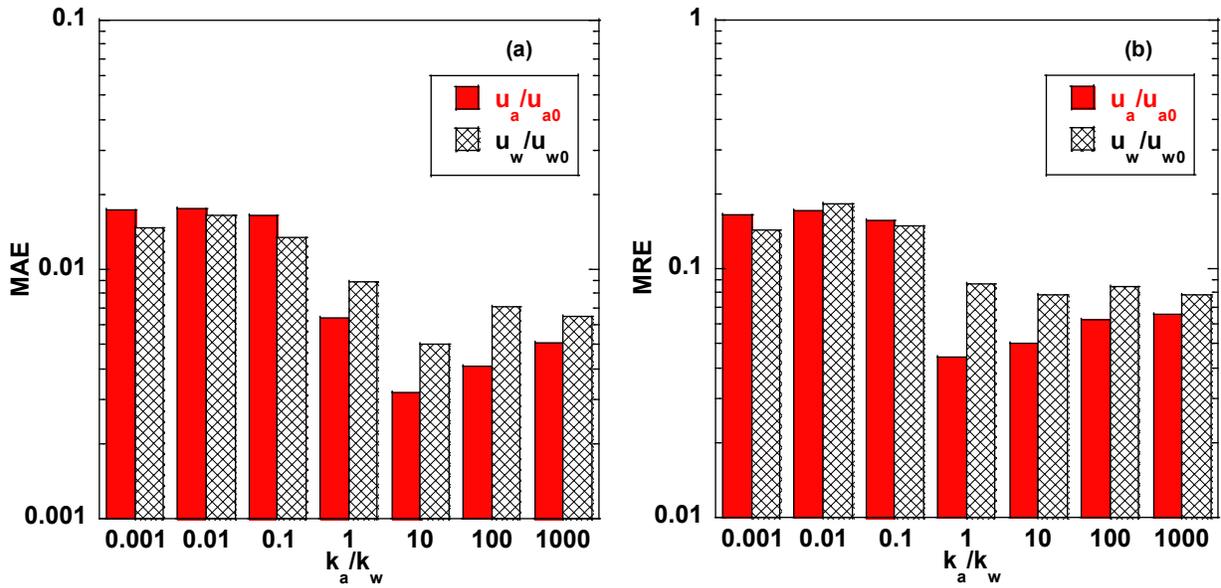

**Fig. 16.** Accuracy vs. air-to-water permeability ratios of $k_a/k_w$ with simplified time segment partition strategy: (a) MAE and (b) MRE for $(u_a/u_{a0})$ and $(u_w/u_{w0})$.

# 6. Summary and Conclusions

This study developed a Lagged Backward-Compatible Physics-Informed Neural Network (LBC-PINN) to simulate one-dimensional unsaturated soil consolidation. The framework introduces normalization of the spatiotemporal domain, lagged compatibility loss, and transfer learning to overcome convergence issues and temporal resolution imbalances in conventional PINNs. Forward predictions were benchmarked against finite-element solutions, and a series of sensitivity and inverse analyses was performed to assess robustness and parameter identifiability. The following conclusions and recommendations are drawn:

1) The LBC-PINN accurately captures the coupled dissipation behavior of pore-air and pore-water pressures over time scales spanning up to $10^{10}$ s. For the recommended segmentation settings (e.g., $N \geq 5$), combined mean absolute errors for $u_a$ and $u_w$ are on the order of $10^{-3} \sim 10^{-2}$, demonstrating excellent agreement with reference FEM solutions. The introduction of lagged compatibility loss between time segments ensures smooth transitions and improved temporal consistency, while transfer learning accelerates convergence across segments.

2) The LBC-PINN framework handles a wide range of permeability ratios ($k_a/k_w$ from 0.001 to 1000), maintaining low error across regimes dominated by either air flow or water flow. A moderate accuracy drop was observed at $k_a/k_w = 0.1$, attributed to strong coupling between air and water phases. Nevertheless, the predictions remained physically consistent throughout the tested range. Sensitivity analyses with respect to network depth, width, and collocation point density show that prediction accuracy improves with increased residual-point sampling, whereas excessively dense boundary sampling yields diminishing returns.

3) The segmented training strategy is essential for resolving the strongly multi-scale temporal response. For a small number of segments, errors remain large and dissipation fronts are under-resolved, whereas segmenting the time domain into four or more logarithmically spaced windows yields stable, high-accuracy solutions. A simplified strategy based on the characteristic air-phase dissipation time ($t_s = H^2/c_{va}$) was proposed to define the initial segment cutoff. The simplified scheme achieves comparable accuracy to full logarithmic segmentation, with absolute errors concentrated only in localized regions (e.g., near the base or at intermediate coupling stages).

The current model is limited to one-dimensional unsaturated soil consolidation. While suitable for many laboratory or idealized field problems, real-world geotechnical systems often involve two- or three-dimensional flow and deformation. Although a physically motivated segmentation strategy was introduced using the characteristic air-phase dissipation time, the segment boundaries and network settings still require manual tuning. Future improvements may involve adaptive segmentation based on dynamic error indicators or automated refinement of training time windows.

# Acknowledgements

The authors declare that this work was carried out independently and did not receive any external funding.

# References

Barden, L., 1965. Consolidation of compacted and unsaturated clays. Geotechnique, 15(3), 267-286.
Bekele, Y.W., 2021. Physics-informed deep learning for one-dimensional consolidation. Journal of Rock Mechanics and Geotechnical Engineering,13(2),420-430.
Biot, M.A., 1941. General theory of three-dimensional consolidation. Journal of applied physics, 12(2), 155-164.
Briaud, J.L., 2023. Geotechnical engineering: unsaturated and saturated soils. John Wiley & Sons.
Cargill, K.W., 1984. Prediction of consolidation of very soft soil. Journal of Geotechnical Engineering, 110(6),775-795.
Cai, S., Mao, Z., Wang, Z., Yin, M. and Karniadakis, G.E., 2021. Physics-informed neural networks (PINNs) for fluid mechanics: A review. Acta Mechanica Sinica, 37(12), 1727-1738.
Chen, Z., Lai, S.K. and Yang, Z., 2024. AT-PINN: Advanced time-marching physics-informed neural network for structural vibration analysis. Thin-Walled Structures, 196, p.111423.
Cheng, Y., Zhang, L.L., Li, J.H., Zhang, L.M., Wang, J.H. and Wang, D.Y., 2017. Consolidation in spatially random unsaturated soils based on coupled flow-deformation simulation. International Journal for Numerical and Analytical Methods in Geomechanics, 41(5), 682-706.
Conte, E., 2004. Consolidation analysis for unsaturated soils. Canadian Geotechnical Journal, 41(4), 599-612.
Cuomo, S., Di Cola, V.S., Giampaolo, F., Rozza, G., Raissi, M. and Piccialli, F., 2022. Scientific machine learning through physics–informed neural networks: Where we are and what's next. Journal of Scientific Computing, 92(3), p.88.
Dubey, S.R., Singh, S.K. and Chaudhuri, B.B., 2022. Activation functions in deep learning: A comprehensive survey and benchmark. *Neurocomputing*, *503*, 92-108.
Esrig, M.I., 1968. Pore pressures, consolidation, and electrokinetics. Journal of the Soil Mechanics and Foundations Division, 94(4),899-921.
Fredlund, D.G. and Hasan, J.U., 1979. One-dimensional consolidation theory: unsaturated soils. Canadian Geotechnical Journal, 16(3), 521-531.
Glorot, X. and Bengio, Y., 2010, March. Understanding the difficulty of training deep feedforward neural networks. In Proceedings of the thirteenth international conference on artificial intelligence and statistics (pp. 249-256). JMLR Workshop and Conference Proceedings.
Goswami, S., Anitescu, C., Chakraborty, S. and Rabczuk, T., 2020. Transfer learning enhanced physics informed neural network for phase-field modeling of fracture. Theoretical and Applied Fracture Mechanics, 106, p.102447.
Gunes Baydin, A., Pearlmutter, B.A., Andreyevich Radul, A. and Siskind, J.M., 2015. Automatic differentiation in machine learning: a survey. arXiv e-prints, pp.arXiv-1502.
He, Q. and Tartakovsky, A.M., 2021. Physics-informed neural network method for forward and backward advection-dispersion equations. Water Resources Research,57(7), p.e2020WR029479.

Ho, L., Fatahi, B. and Khabbaz, H., 2014. Analytical solution for one-dimensional consolidation of unsaturated soils using eigenfunction expansion method. International Journal for Numerical and Analytical Methods in Geomechanics, 38(10), 1058-1077.

Ho, L., Fatahi, B., 2016. One-dimensional consolidation analysis of unsaturated soils subjected to time-dependent loading. International Journal of Geomechanics, 16(2), 04015052.

Ho, L. and Fatahi, B., 2018. Analytical solution to axisymmetric consolidation of unsaturated soil stratum under equal strain condition incorporating smear effects. International Journal for Numerical and Analytical Methods in Geomechanics, 42(15), 1890-1913.

Huang, J., Griffiths, D.V. and Fenton, G.A., 2010. Probabilistic analysis of coupled soil consolidation. Journal of Geotechnical and Geoenvironmental Engineering, 136(3), 417-430.

Jagtap, A.D. and Karniadakis, G.E., 2020. Extended physics-informed neural networks (XPINNs): A generalized space-time domain decomposition based deep learning framework for nonlinear partial differential equations. Communications in Computational Physics, 28(5).

Kiyani, E., Shukla, K., Urbán, J.F., Darbon, J. and Karniadakis, G.E., 2025. Which Optimizer Works Best for Physics-Informed Neural Networks and Kolmogorov-Arnold Networks?. *arXiv preprint arXiv:2501.16371*.

Kumar, G.S. and Premalatha, K., 2021. Securing private information by data perturbation using statistical transformation with three dimensional shearing. Applied Soft Computing, 112, p.107819.

Lan, P., Su, J.J., Ma, X.Y. and Zhang, S., 2024. Application of improved physics-informed neural networks for nonlinear consolidation problems with continuous drainage boundary conditions. Acta Geotechnica, 19(1), 495-508.

Lawal, Z.K., Yassin, H., Lai, D.T.C. and Che Idris, A., 2022. Physics-informed neural network (PINN) evolution and beyond: A systematic literature review and bibliometric analysis. Big Data and Cognitive Computing, 6(4), p.140.

Lloret, A. and Alonso, E.E., 1980. Consolidation of unsaturated soils including swelling and collapse behaviour. Géotechnique, 30(4), 449-477.

Liu, E., Yu, H.S., Deng, G., Zhang, J. and He, S., 2014. Numerical analysis of seepage–deformation in unsaturated soils. Acta Geotechnica, 9(6), 1045-1058.

Liu, Y., Liu, W., Yan, X., Guo, S. and Zhang, C.A., 2023. Adaptive transfer learning for PINN. Journal of Computational Physics, 490, p.112291.

Lu, L., Meng, X., Mao, Z. and Karniadakis, G.E., 2021. DeepXDE: A deep learning library for solving differential equations. SIAM review, 63(1), pp.208-228.

Mao, Z. and Meng, X., 2023. Physics-informed neural networks with residual/gradient-based adaptive sampling methods for solving partial differential equations with sharp solutions. Applied Mathematics and Mechanics, 44(7), 1069-1084.

Mattey, R. and Ghosh, S., 2022. A novel sequential method to train physics informed neural networks for Allen Cahn and Cahn Hilliard equations. Computer Methods in Applied Mechanics and Engineering, 390, p.114474.

Mesri, G. and Rokhsar, A., 1974. Theory of consolidation for clays. Journal of the Geotechnical Engineering Division, 100(8), 889-904.

Qin, A.F., Chen, G.J., Tan, Y.W. and Sun, D.A., 2008. Analytical solution to one-dimensional consolidation in unsaturated soils. Applied mathematics and mechanics, 29(10), 1329-1340.

Qin, A., Sun, D.A. and Tan, Y., 2010. Analytical solution to one-dimensional consolidation in unsaturated soils under loading varying exponentially with time. Computers and Geotechnics, 37(1-2), 233-238.

Reddy, J.N., 1993. An introduction to the finite element method. New York, 27(14).

Rahaman, N., Baratin, A., Arpit, D., Draxler, F., Lin, M., Hamprecht, F., Bengio, Y. Courville, A., 2019, May. On the spectral bias of neural networks. In International conference on machine learning (5301-5310). PMLR.

Raissi, M., Perdikaris, P. and Karniadakis, G.E., 2019. Physics-informed neural networks: A deep learning framework for solving forward and inverse problems involving nonlinear partial differential equations. Journal of Computational physics, 378, 686-707.

Scott, R.F., 1963. Principles of Soil Mechanics, Addison-Wesley. Reading, Massachusetts, p.54.

Shan, Z., Ling, D. and Ding, H., 2012. Exact solutions for one-dimensional consolidation of single-layer unsaturated soil. International Journal for Numerical and Analytical Methods in Geomechanics, 36(6), 708-722.

Sharma, P., Malik, N., Akhtar, N. and Rohilla, H., 2013. feedforward neural network: A Review. International Journal of Advanced Research in Engineering and Applied Sciences (IJAREAS), 2(10), 25-34.

Tartakovsky, A.M., Marrero, C.O., Perdikaris, P., Tartakovsky, G.D. and Barajas-Solano, D., 2020. Physics-informed deep neural networks for learning parameters and constitutive relationships in subsurface flow problems. Water Resources Research, 56(5), p.e2019WR026731.

Tang, Y., Taiebat, H.A. and Russell, A.R., 2018. Numerical modeling of consolidation of unsaturated soils considering hydraulic hysteresis. International Journal of Geomechanics, 18(2), p.04017136.

Terzaghi, K., 1943. Theory of consolidation. Theoretical soil mechanics, 265-296.

Toscano, J.D., Oommen, V., Varghese, A.J., Zou, Z., Ahmadi Daryakenari, N., Wu, C. and Karniadakis, G.E., 2025. From pinns to pikans: Recent advances in physics-informed machine learning. *Machine Learning for Computational Science and Engineering*, *1*(1), 1-43.

Xu, S., Dai, Y., Yan, C., Sun, Z., Huang, R., Guo, D. and Yang, G., 2025. On the preprocessing of physics-informed neural networks: How to better utilize data in fluid mechanics. Journal of Computational Physics, 528, p.113837.

Wang, J., Mo, Y.L., Izzuddin, B. and Kim, C.W., 2023. Exact Dirichlet boundary physics-informed neural network EPINN for solid mechanics. Computer Methods in Applied Mechanics and Engineering, 414, p.116184.

Wang, L., Xu, Y., Xia, X., Li, L. and He, Y., 2020. A series of semianalytical solutions of one-dimensional consolidation in unsaturated soils. International Journal of Geomechanics, 20(6), 06020005.

Wang, S., Teng, Y. and Perdikaris, P., 2021. Understanding and mitigating gradient flow pathologies in physics-informed neural networks. *SIAM Journal on Scientific Computing*, *43*(5), pp.A3055-A3081.

Wang, S., Yu, X. and Perdikaris, P., 2022. When and why PINNs fail to train: A neural tangent kernel perspective. Journal of Computational Physics, 449, 110768.

Wang, Y., Shi, C., Shi, J. and Lu, H., 2024. Data-driven forward and inverse analysis of two-dimensional soil consolidation using physics-informed neural network. Acta Geotechnica,19 (12), 8051-8069.

Wong, T.T., Fredlund, D.G. and Krahn, J., 1998. A numerical study of coupled consolidation in unsaturated soils. Canadian Geotechnical Journal, 35(6), 926-937.

Zhang, P., Yin, Z.Y. and Sheil, B., 2024. A physics-informed data-driven approach for consolidation analysis. Géotechnique, 74(7), 620-631.

Zhou, W.H., Zhao, L.S. and Li, X.B., 2014. A simple analytical solution to one-dimensional consolidation for unsaturated soils. International Journal for Numerical and Analytical Methods in Geomechanics, 38(8), 794-810.

Zhou, W.H., Zhao, L.S., Garg, A. and Yuen, K.V., 2017. Generalized analytical solution for the consolidation of unsaturated soil under partially permeable boundary conditions. International Journal of Geomechanics, 17(9), .04017048.

Zhou, W.H., 2013. Axisymmetric consolidation of unsaturated soils by differential quadrature method. Mathematical Problems in Engineering, 2013(1), p.497161.

Zhou, W.H. and Zhao, L.S., 2014. Consolidation of a two-layer system for unsaturated soil with the differential quadrature method. In Geo-Congress 2014: Geo-Characterization and Modeling for Sustainability (pp. 3994-4003).

Zhou, H., Wu, H., Sheil, B. and Wang, Z., 2025. A self-adaptive physics-informed neural networks method for large strain consolidation analysis. Computers and Geotechnics, 181, p.107131.